\documentclass[pdflatex,sn-mathphys-num]{sn-jnl}

\usepackage[utf8]{inputenc}
\usepackage{textcomp}
\usepackage{algorithm,algpseudocode}
\usepackage{graphicx}
\usepackage{amsmath}
\usepackage[version=4]{mhchem}
\usepackage{siunitx}
\usepackage{longtable,tabularx}

\usepackage{makecell} 

\usepackage[inkscapeformat=png]{svg}
\usepackage{mathtools}
\usepackage{gensymb}
\usepackage{float}
\usepackage{graphicx}
\usepackage{xcolor}
\usepackage{tabularx}
\usepackage{adjustbox}
\usepackage{array}
\usepackage{threeparttable}
\usepackage{float}

\usepackage{graphicx}%
\usepackage{multirow}%
\usepackage{amsmath,amssymb,amsfonts}%
\usepackage{amsthm}%
\usepackage{mathrsfs}%
\usepackage[title]{appendix}%
\usepackage{xcolor}%
\usepackage{textcomp}%
\usepackage{manyfoot}%
\usepackage{booktabs}%
\usepackage{algorithm}%
\usepackage{algorithmicx}%
\usepackage{algpseudocode}%
\usepackage{listings}%

\theoremstyle{thmstyleone}%
\theoremstyle{thmstyletwo}%

\theoremstyle{thmstylethree}%

\begin{document}

\title[Adaptive Relative Pose Estimation Framework with Dual Noise Tuning for Uncooperative Target]{Adaptive Relative Pose Estimation Framework with Dual Noise Tuning for Uncooperative Target}


\author*[1]{\fnm{Batu} \sur{Candan}}\email{dukynuke@iastate.edu}

\author[2]{\fnm{Murat Berke} \sur{Oktay}}\email{berke.oktay@metu.edu.tr}

\author[3]{\fnm{Simone} \sur{Servadio}}\email{servadio@iastate.edu}

\affil*[1]{\orgdiv{Aerospace Engineering Department}, \orgname{Iowa State University}, \orgaddress{\city{Ames}, \postcode{50011}, \state{Iowa}, \country{USA}}}

\affil[2]{\orgdiv{Mechanical Engineering Department}, \orgname{Middle East Technical University}, \orgaddress{\city{Ankara}, \postcode{06800}, \country{Türkiye}}}

\affil[3]{\orgdiv{Aerospace Engineering Department}, \orgname{Iowa State University}, \orgaddress{\city{Ames}, \postcode{50011}, \state{Iowa}, \country{USA}}}


\abstract{Accurate and robust relative pose estimation is crucial for enabling challenging Active Debris Removal (ADR) missions targeting derelict satellites such as ESA's ENVISAT. This work presents a complete pipeline integrating computer vision techniques with adaptive nonlinear filtering to address this challenge. A Convolutional Neural Network (CNN) detects corners from chaser camera, whose 2D coordinates are converted to 3D measurements using camera modeling. These measurements are fused within an Unscented Kalman Filter (UKF) framework, selected for its ability to handle nonlinear relative dynamics, to estimate the full relative pose. Key contributions include the integrated system architecture and a dual adaptive strategy within the UKF: dynamic tuning of the measurement noise covariance compensates for varying CNN measurement uncertainty, while adaptive tuning of the process noise covariance, utilizing measurement residual analysis, accounts for unmodeled dynamics or maneuvers online. This dual adaptation enhances robustness against both measurement imperfections and dynamic model uncertainties. The performance of the proposed adaptive integrated system is evaluated through high-fidelity simulations using a realistic ENVISAT model, comparing estimates against ground truth under various conditions, including measurement outages. This comprehensive approach offers an enhanced solution for robust onboard relative navigation, significantly advancing the capabilities required for safe proximity operations during ADR missions.}




\maketitle

\section{Introduction}\label{sec1}

The capability to estimate the relative pose of uncooperative targets, such as derelict satellites, is critical for enabling future ADR, on-orbit servicing, and space situational awareness missions. These scenarios are characterized by severe visibility challenges, unpredictable dynamics, and uncertain sensor performance, all of which make traditional, rigid estimation pipelines inadequate. In particular, ESA’s ENVISAT has become a key benchmark target for ADR due to its large size, complex structure, and non-cooperative nature \cite{ESTABLE202052, envisatdl, cakal}. Moreover, it represents a risk for the sustainability of the LEO space environment \cite{servadio2024risk,servadio2024threat}. Recent advances in vision-based navigation and machine learning have dramatically improved object detection in space imagery \cite{lentaris, Rizzuto2025, furano, RENAUT2025231}. CNNs and deep learning models enable the detection of structural markers on a spacecraft from camera images, even under challenging lighting and motion conditions \cite{zhou, BECHINI202420, DIXON2025110055}. However, these visual outputs still face significant uncertainty due to sensor limitations, jitter, and partial or complete occlusions \cite{Gao2023}. Therefore, fusing these noisy, intermittent detections into a reliable navigation solution remains a central challenge. To address this, Kalman filtering and its nonlinear variants, particularly the UKF, have emerged as robust solutions for visual navigation \cite{WEI2024108832, oproukf, servadiojsr, aukff, servadio2020recursive, servadio2020nonlinear}. Yet, traditional filters assume fixed noise models, which often do not reflect the real-world variability encountered during proximity operations \cite{Driedger2020,kaidan,sampath}. This results in either over-conservative or overly confident estimations depending on operational conditions. While prior works have attempted to tune the process noise covariance matrix $\mathbf{Q}$ adaptively, many approaches suffer from either computational complexity or limited scalability. Mamich et al. \cite{mamich} propose a variational Bayesian approach to estimate $\mathbf{Q}$ using a probabilistic inference framework, but this requires iterative updates of the evidence lower bound, posing a significant challenge for real-time systems. Similarly, Moghe et al. \cite{moghe} explore dynamic $\mathbf{Q}$ estimation in reinforcement learning settings, relying on sampling-based uncertainty propagation that increases computational burden with state dimension. Zanetti and D’Souza \cite{zanetti} analyze filter robustness under observability constraints and propose cautious tuning heuristics for $\mathbf{Q}$, but lack an adaptive mechanism that reacts online to system dynamics. 

In contrast to existing approaches, this paper introduces a fully adaptive, dual-noise tuning framework for vision-based relative navigation, with particular emphasis on spacecraft proximity operations under uncertainty. Our main novelty lies in the joint online adaptation of both the process noise covariance matrix $\mathbf{Q}$ and the measurement noise covariance matrix $\mathbf{R}$ using real-time filter statistics, without requiring prior eclipse scheduling, tuning heuristics, or batch post-processing. The $\mathbf{Q}$ adaptation leverages a forward–backward residual matrix, inspired by the Rauch–Tung–Striebel (RTS) smoother \cite{Sarkka2013}, to capture the mismatch between prior and propagated sigma points and to adjust process noise during unobservable or eclipse phases. Simultaneously, the $\mathbf{R}$ tuning strategy employs a residual consistency filter that tracks innovation growth in real-time and injects per-marker correction factors through the Multiple Tuning Factor (MTF) diagonal matrix inspired by the author's previous works and others \cite{batumtf, sokencandan, sokenukf, candan_2022}. This dual-adaptation architecture allows the filter to remain responsive to both system-driven and measurement-driven uncertainty, achieving consistent and bounded covariance behavior. To the best of our knowledge, this is the first integrated application of RTS-inspired $\mathbf{Q}$ adaptation with innovation-based $\mathbf{R}$ tuning for monocular vision-based relative navigation in space. Moreover, as a final contribution, we introduce a synthetic LiDAR sensor model, implemented in Blender, to complement the vision-based measurements. Using a ray-casting procedure aligned with the camera geometry, the LiDAR provides dense depth samples co-registered to the image plane. While this improves measurement diversity, practical mismatches arise between LiDAR points and projected corner locations, leading to systematic offsets in the depth channel. To address this, we augment the filter states with a LiDAR depth bias variable and developed a measurement fusion scheme that associates LiDAR points with corner detections. This implementation significantly enhances the robustness of the filter by accounting for real-world sensor imperfections and maintaining consistent estimation performance.  The framework's efficacy is demonstrated on high-fidelity simulations of ESA's ENVISAT, where our approach outperforms previous work and variational Bayesian methods in both accuracy and computational efficiency, particularly under full measurement outages. The results from this effort advance the need for ADR \cite{simha2025optimal} for the safety of the space environment and avoiding the predicted Kessler's syndrome \cite{jang2025new}.

The paper is structured as follows. Section~\ref{sec:methodology} presents the overall methodology, including the spacecraft dynamics, simulation framework, and data generation pipeline. Section~\ref{sec:cnn} details the CNN model used for marker detection and the preprocessing steps applied to enhance robustness. Section~\ref{sec:filtering} describes the UKF formulation, including sigma point generation, prediction, and correction stages. Section~\ref{sec:measurement} introduces the measurement model, highlighting the corner association and the statistical derivation of the measurement noise covariance. Section~\ref{sec:adaptation} develops the dual adaptive strategy for process and measurement noise tuning, combining innovation-based residual filtering with a smoothing-inspired covariance update. Section~\ref{sec:results} evaluates the framework through extensive Monte Carlo simulations under eclipse and non-eclipse conditions. Section~\ref{sec:meas_improvement} discusses the LiDAR-augmented measurement model and bias correction strategy as an improvement to a measurement model, while Section~\ref{sec:conclusions} summarizes the findings and outlines future research directions.

\section{Methodology}
\label{sec:methodology}
This section details the proposed methodology, covering spacecraft dynamics (chaser and target), the ENVISAT simulation parameters, and the data preparation steps for image processing and estimation. Note that the details of these derivations and steps are discussed in detail in \cite{candanMdpi, phdthesis}.

\subsection{Dynamics}
\subsubsection{Absolute Chaser Motion:}
To model the chaser's motion, the following equations are used. 

\begin{align}
   \ddot{\Bar{r}} = \Bar{r} \dot{\theta}^2 - \frac{\mu}{\Bar{r}^2} \\
   \ddot{\theta} = -2 \frac{\dot{\Bar{r}} \dot{\theta}}{\Bar{r}}
\end{align}
They incorporate Earth's gravitational parameter ($\mu$), the chaser's position vector with respect to Earth ($\bar{r}$), and its orbital true anomaly ($\theta$).

\subsubsection{Relative Translational Dynamics:}
The equations describing ENVISAT's relative translational motion are derived using the chaser's Local-Vertical-Local-Horizontal frame as the reference. Within this frame, the target's relative position is represented by the vector $\mathbf{r}_{\textbf{r}}$ and its relative velocity by $\mathbf{v}_{\textbf{r}}$, as defined in:

\begin{align}
    \mathbf{r}_\textbf{r} &= x\hat{\mathbf{i}}+y\hat{\mathbf{j}}+z\hat{\mathbf{k}} \\
   \mathbf{v}_\textbf{r} &= \Dot{x}\hat{\mathbf{i}}+\Dot{y}\hat{\mathbf{j}}+\Dot{z}\hat{\mathbf{k}} 
\end{align}
Here, \(x\), \(y\), and \(z\) are the elements of the position vector \( \mathbf{r}_\textbf{r} \) within the chaser frame, and \( \hat{\mathbf{i}} \), \( \hat{\mathbf{j}} \), and \( \hat{\mathbf{k}} \) represent the unit vectors of the reference frame. So, the equations of relative motion of the target can be written in the following way:
\begin{align}
    \ddot{x} = 2\dot{\theta}\dot{y} + \ddot{\theta}y + \dot{\theta}^2x - \frac{\mu(\ddot{r}+x)}{[(\Bar{r}+x)^2+y^2+z^2]^{3/2}} + \frac{\mu}{\Bar{r}^2}
\\
    \ddot{y} = -2\dot{\theta}\dot{x} - \ddot{\theta}x + \dot{\theta}^2y - \frac{\mu y}{[(\Bar{r}+x)^2+y^2+z^2]^{3/2}}
\\
    \ddot{z} = -\frac{\mu z}{[(\Bar{r}+x)^2+y^2+z^2]^{3/2}}
\end{align}

\subsubsection{Rotational Dynamics:}
Let $\mathbf{\Gamma}$ be the rotation matrix defining the relative orientation between the chaser and target body-fixed frames. The relative angular velocity $\boldsymbol{\omega}_r$, expressed in the target's body frame, depends on the individual body-frame angular velocities of the chaser ($\boldsymbol{\omega}_c$) and the target ($\boldsymbol{\omega}_t$).
\begin{align}
    \mathbf{\omega_r} = \mathbf{\omega_t} - \mathbf{\Gamma} \mathbf{\omega_c}
\\
    \dot{\mathbf{\omega}}_\textbf{r} = \dot{\mathbf{\omega}}_\textbf{t} - \mathbf{\Gamma} \dot{\mathbf{\omega}}_\textbf{c} + \mathbf{\omega_r} \wedge \mathbf{\Gamma} \mathbf{\omega_c}
\end{align}
Modified Rodrigues Parameters (MRPs) are utilized to parametrize the relative attitude, represented by the rotation matrix $\mathbf{\Gamma}$. This choice leverages the MRPs' three-parameter and singularity-free nature for describing rotation. While quaternions ($\mathbf{q}$, including vector components $q_1, q_2, q_3,$ and scalar $q_4$) also represent orientation, MRPs are specifically utilized for the attitude parametrization in this work. The satellite's angular velocity vector is denoted $\boldsymbol{\omega}$, with components $\omega_x, \omega_y,$ and $\omega_z$.
\begin{align}
    \mathbf{q} = \begin{bmatrix}
                    \Bar{\mathbf{q}} \\
                    q_4
                \end{bmatrix}
\\
    \Bar{\mathbf{q}} = \begin{bmatrix}
                        q_1,
                        q_2,
                        q_3
                        \end{bmatrix}^T = \mathbf{\hat{n}} \sin{\frac{\phi}{2}}
\\
    q_4 = \cos{\frac{\phi}{2}}
\end{align}
Defining $\mathbf{\hat{n}}$ as the unit vector along the rotation axis and $\phi$ as the corresponding rotation angle, the relationship connecting MRPs to quaternions is given as follows:
\begin{equation}
    \mathbf{p} = \frac{\Bar{\mathbf{q}}}{(1+q_4)} = \mathbf{\hat{n}} \tan{\frac{\phi}{4}}
\end{equation}
where $\mathbf{p}$ is the 3D MRP vector. The kinematic equation of motion can be derived from the target's relative angular velocity. This relationship yields the following expression for the time derivative of the MRPs:
\begin{equation}
    \dot{\mathbf{p}} = \frac{1}{4}\left[(1-\mathbf{p}^T\mathbf{p})\mathbf{I}_3 + 2\mathbf{p}\mathbf{p}^T + 2[\mathbf{p}\wedge]\right]\mathbf{\omega_r}
\end{equation}
Note that $\mathbf{I}_3$ is a $3 \times 3$ identity matrix, and $[\mathbf{p}\wedge]$ identifies the $3 \times 3$ cross product matrix given as,
\begin{equation}
    [\mathbf{p}\wedge] = \begin{bmatrix}
                        0 & -p_3 & p_2 \\
                        p_2 & 0 & -p_1 \\
                        -p_2 & p_1 & 0
                        \end{bmatrix}
\end{equation}
Therefore, the rotation matrix $\mathbf{\Gamma}$, which relates the body-fixed frame of the chaser to that of the target, can be expressed as follows:
\begin{align}
    \epsilon_1 = 4 \frac{1-\mathbf{p}^T\mathbf{p}}{(1+\mathbf{p}^T\mathbf{p})^2}
\\
    \epsilon_2 = 8 \frac{1}{(1+\mathbf{p}^T\mathbf{p})^2}
\\
    \mathbf{\Gamma}(\mathbf{p}) = \mathbf{I}_3 - \epsilon_1 [\mathbf{p}\wedge] + \epsilon_2 [\mathbf{p}\wedge]^2
\end{align}

While the chaser's absolute rotational dynamics follow the torque-free Euler equations, the relative attitude dynamics are obtained by substituting kinematic relationships into the target's absolute Euler equations.
\begin{equation}
    \mathbf{J}_\textbf{t} \dot{\omega}_\textbf{r} + \omega_\textbf{r} \wedge \mathbf{J}_\textbf{t} \omega_\textbf{r} = \mathbf{M}_{app} -\mathbf{M}_{g} - \mathbf{M}_{ci}
\end{equation}
Here, $\mathbf{J}_{\textbf{t}}$ is the inertia matrix of the target spacecraft, while $\mathbf{M}_{\text{app}}$, $\mathbf{M}_{\text{g}}$, and $\mathbf{M}_{\text{ci}}$ represent the apparent, gyroscopic, and chaser-inertial torques, respectively. The specific definitions for these terms are given as:
\begin{align}
    \mathbf{M}_{app} = \mathbf{J}_\textbf{t} \omega_\textbf{r} \wedge \mathbf{\Gamma} \omega_\textbf{c}
\\
    \mathbf{M}_{g} = \mathbf{\Gamma} \omega_\textbf{c} \wedge \mathbf{J}_\textbf{t} \mathbf{\Gamma} \omega_\textbf{c} + \omega_\textbf{r} \wedge \mathbf{J}_\textbf{t} \omega_\textbf{c} + \mathbf{\Gamma} \omega_\textbf{c} \wedge \mathbf{J}_\textbf{t} \omega_\textbf{r}
\\
    \mathbf{M}_{ci} = \mathbf{J}_\textbf{t} \mathbf{\Gamma} \dot{\omega}_\textbf{c}
\end{align}

\subsection{ENVISAT Satellite Simulation}
\begin{figure}[H]
    \makebox[\textwidth][c]{\includegraphics[width=\textwidth]{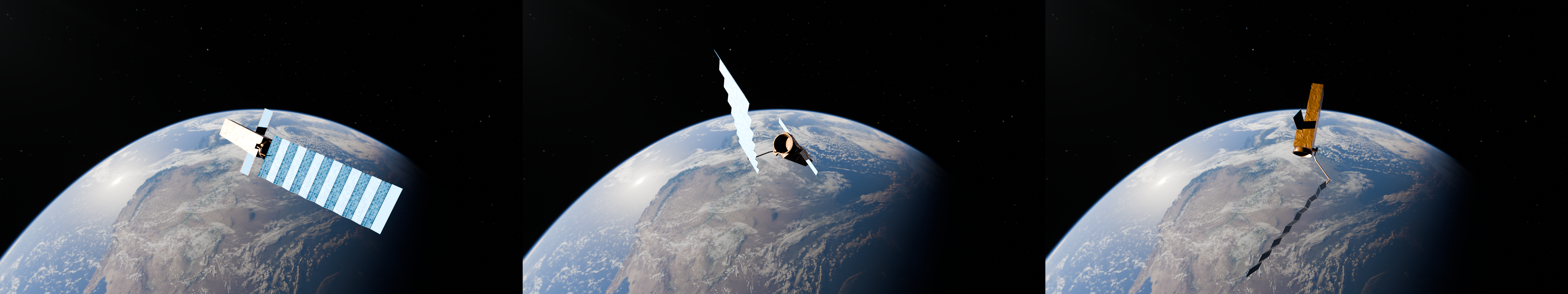}}
    \caption{Simulated images of ENVISAT satellite}
    \centering
    \label{fig:mergedsim}
\end{figure}

To generate a realistic dataset reflecting space conditions, a simulation environment for the ENVISAT satellite was established and implemented using Python in Blender environment, which is a crucial novelty unlike previous works \cite{candan, candanMdpi}. This implementation includes a custom pinhole camera model and projection logic, also coded in Python, based on specified intrinsic parameters, leading to a defined camera matrix. The simulated 3D model, derived from ESA-provided data, focuses on the satellite's main rectangular body, specifically to assess corner detection algorithms \cite{DiLizia2017Assessment}. The Blender simulation generates synthetic images by projecting the 3D model onto the image plane according to its time-varying state using the custom camera projection routines. A sequence of these generated images is demonstrated in Fig.~\ref{fig:mergedsim}. For each simulation step, key data were stored: the 12-component relative state vector (relative position, velocity, attitude represented by MRPs, and angular velocity) between the target and chaser, along with the true 2D pixel coordinates and depth values for each of the main corners as projected onto the image. These pixel locations and depths, calculated and saved by the Python script, serve as the precise ground truth for validating the corner detection algorithm. 

In addition to geometry and camera modeling, realistic illumination was incorporated into the Blender environment \footnote{Blender 4.2.0 software is used in this work.}{}. A directional sunlight source was introduced to approximate solar illumination in orbit, with irradiance scaled to the solar constant ($1361~\text{W/m}^2$) and oriented relative to the scene to cast sharp shadows on the satellite body. The light was positioned at a large virtual distance to replicate parallel rays. This lighting setup produces high-contrast shading and realistic occlusion patterns on the ENVISAT model, which in turn improves the fidelity of the synthetic images. The simulation was run at 24~fps (though 1~Hz images are extracted for Monte Carlo simulations from the chaser's onboard camera) to maintain compatibility with Blender’s rendering pipeline. Each frame stores not only the rendered image but also the projected corner positions, their depths, and visibility flags (via ray-casting checks, which will be discussed in the following section), ensuring a consistent ground truth dataset. Note that until this point, in recent works \cite{candanMdpi, candan}, the camera simulation was conducted inefficiently in MATLAB, and then in Python, while proper camera transformation, dynamic simulation, and rendering are implemented in Blender now \cite{candan, candanMdpi}. Table \ref{tab:camera_properties} shows the camera parameters selected in this study. The decision to use these parameters has been taken after reviewing the related works and the literature \cite{pauly, pauly2}.

\begin{table}[htbp] 
    \centering 
    \caption{Camera intrinsic properties and simulation parameters}
    \label{tab:camera_properties}
    \begin{tabular}{|l|l|} 
        \hline
        \textbf{Parameter} & \textbf{Value} \\
        \hline
        $FoV$ (Field of view) & 45$\degree$ \\
        \hline
        $f_x$ (Focal length in x-direction) & 1920 (pxs) \\
        \hline
        $f_y$ (Focal length in y-direction) & 1280 (pxs) \\
        \hline
        $c_x$ (Principal point x-coordinate) & 960 (pxs) \\
        \hline
        $c_y$ (Principal point y-coordinate) & 640 (pxs) \\
        \hline
        Affine translation (std. dev.) & 3 pixels (in both $x$ and $y$) \\
        \hline
        Affine rotation (std. dev.) & 1$\degree$ \\
        \hline
        Additive Gaussian noise (std. dev.) & 0.001 (normalized intensity) \\
        \hline
    \end{tabular}
\end{table}

\section{Convolutional Neural Network}
\label{sec:cnn}
\begin{figure}[H]
\centering
    {\includegraphics[width=0.85\textwidth]{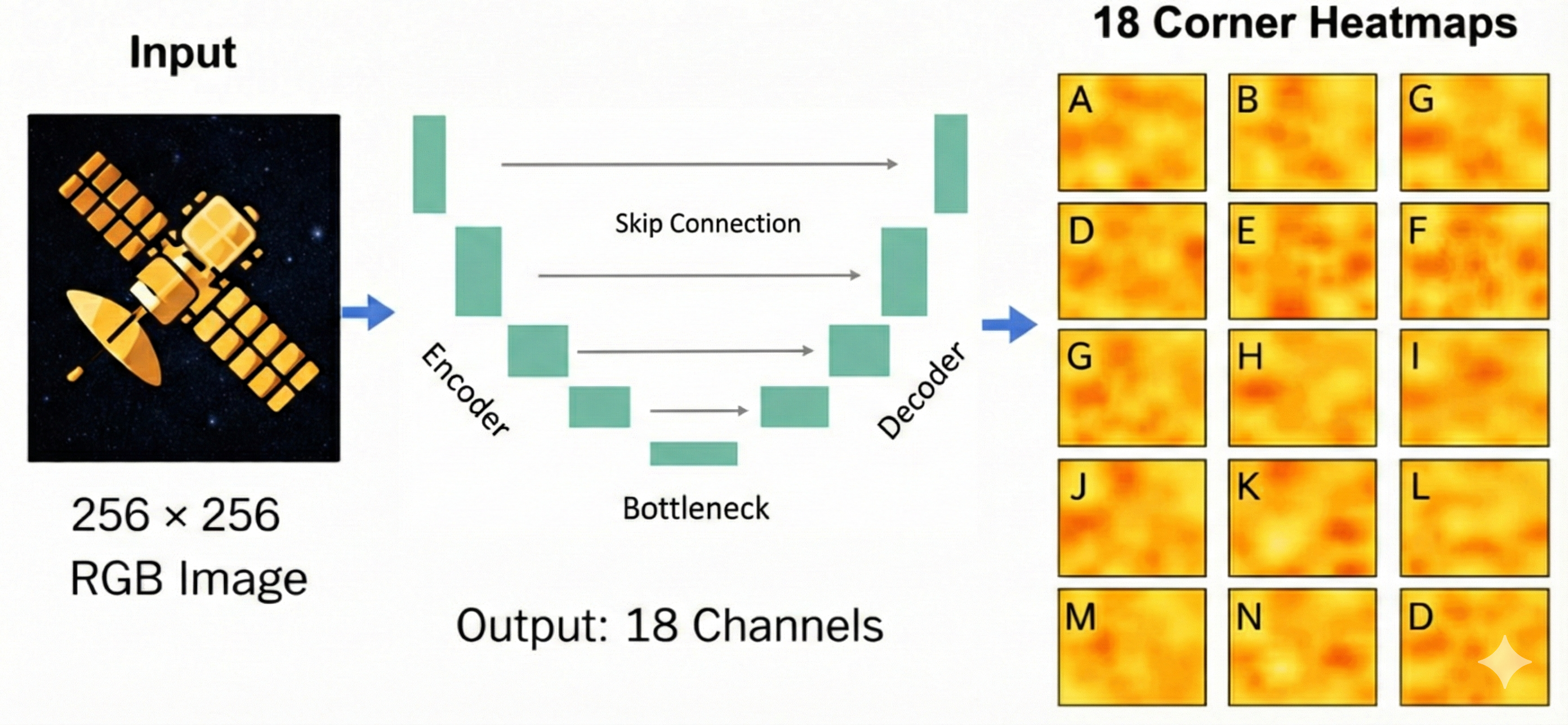}}
    \caption{Overview of TinyCornerNET architecture}
    \label{fig:tinycornernet}
\end{figure}

CNNs are deep learning models that can process grid-like data, such as images, automatically learning spatial feature hierarchies using convolutional, and fully connected layers \cite{alexnet, hrnet}. Compared to other neural networks, CNNs generally require fewer trainable parameters, enhancing generalization and reducing overfitting. CNN architectures have evolved significantly in recent years, often increasing depth and employing specialized blocks to enhance performance in various applications \cite{alzubaidi-2021}. In this project, the chosen neural network model is our lightweight CNN architecture design, named TinyCornerNET, developed specifically for corner detection on the ENVISAT satellite \cite{candanMdpi}. It adopts a heatmap-based keypoint detector implemented as a U-Net++ encoder-decoder network with a ResNet-34 backbone \cite{unet, unetplus}. In contrast to more computationally demanding models such as L-CNN \cite{lcnn} or other networks \cite{lotti, piazza, Rathinam2020, chenDL, heatmap1}, TinyCornerNET is tailored for simplicity and efficiency, making it ideal for deployment in real-time, resource-constrained spaceborne environments. The network architecture comprises a sequence of convolutional layers with ReLU activations, followed by a final convolutional layer with a sigmoid activation to generate a corner likelihood map. The complete network structure is illustrated in Figure~\ref{fig:tinycornernet}. A dataset of 6035 Blender-rendered images is used for training, where each image is accompanied by a CSV file containing the projected pixel coordinates of the labeled markers. Input images are normalized using the ImageNet mean and standard deviation to match the statistics expected by the ResNet-34 encoder. We train the network using a heatmap-based focal loss. This loss drives the model to produce a strong response at the true corner location while down-weighting the overwhelming number of background pixels, which helps address the severe imbalance between corner and non-corner pixels. In addition, only markers that are visible contribute to the loss, so the network is not penalized for occluded corners. Optimization is performed using AdamW, which is Adam with decoupled weight decay and typically provides stable convergence for encoder-decoder architectures \cite{adamw}. We use a learning rate of $1\times10^{-4}$ with a batch size of 16, and train the network for 80 epochs.

\begin{figure}[H]
    \makebox[\textwidth][c]{\includegraphics[width=\textwidth]{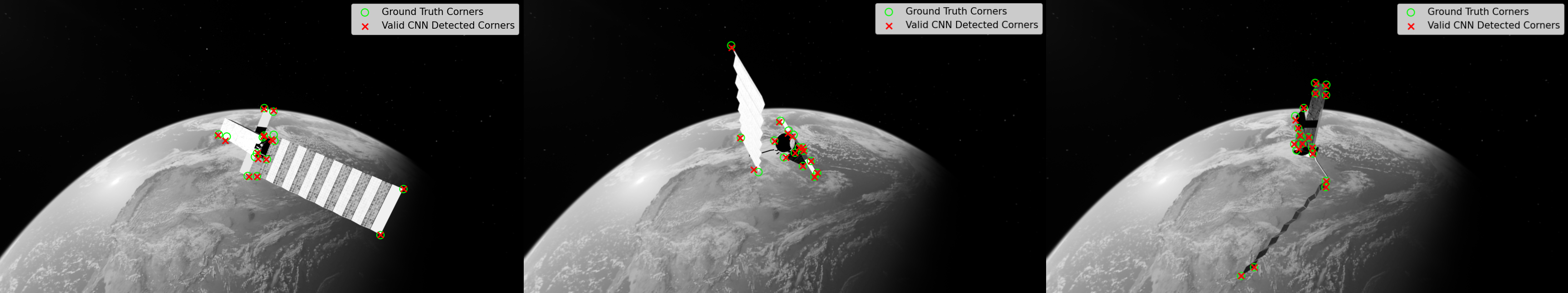}}
    \caption{Preliminary results of CNN algorithm}
    \centering
    \label{fig:markerAsso2}
\end{figure}

Fig. \ref{fig:markerAsso2} shows how effective the CNN algorithm is in detecting the corners of satellites in different orientations and illumination conditions within a simulated space environment. It also indicates that the marker association algorithm, which will be discussed in the next sections, works efficiently and is able to match the detected corners with the specific ground truth corners, which are labeled and known.
Note that even though most of the corners can be projected into the image frame, some of them are not actually visible to the camera due to self-occlusion or obstruction by other satellite components. To ensure the realism and accuracy of the generated ground-truth data, we perform a visibility check using Blender's raycasting functionality. Specifically, each corner's 3D position is tested by casting a ray from the camera toward the marker to determine whether it is directly observable or blocked by other geometry in the scene. Only markers that pass this visibility test are included in the final dataset for projection and depth association. This approach improves the fidelity of our simulation by mimicking realistic line-of-sight constraints in space-borne imaging scenarios.

\section{Filtering}
\label{sec:filtering}
UKF is selected to estimate the relative pose of ENVISAT, due to its capacity to process nonlinear systems. Unlike the Extended Kalman Filter (EKF), which linearizes system dynamics, the UKF applies the unscented transformation to directly process nonlinearity. The approach propagates chosen sigma points to capture state uncertainty, while significantly retaining a linear measurement update suitable for onboard computational constraints \cite{cavenagoDA}.

\subsection{The Filtering Algorithm}
UKF sigma point weights are determined by parameters $\alpha, \beta, \kappa$, which control the spread and scaling of the points. The scaling parameter $\lambda$ is computed as:
\begin{equation}
\lambda = \alpha^2 (L + \kappa) - L
\end{equation}
where $L=12$ is the state vector dimension. The weights for calculating the mean ($W_m$) and covariance ($W_c$) from the sigma points are:

\begin{align}
W_m^{(0)} &= \frac{\lambda}{L + \lambda} \\
W_c^{(0)} &= \frac{\lambda}{L + \lambda} + (1 - \alpha^2 + \beta) \\
W_m^{(i)} = W_c^{(i)} &= \frac{1}{2(L + \lambda)}, \quad i = 1, \ldots, 2L
\end{align}

The prediction step of the filter starts with the generation of well-chosen sigma points.
Sigma points $\chi_{k-1}^{(i)}$ are generated around the previous state estimate $\hat{\mathbf{x}}_{k-1}$ using the covariance $\mathbf{P}_{k-1}$:
\begin{align}
\chi_{k-1}^{(i)} = 
\begin{cases}
  \hat{\mathbf{x}}_{k-1} & \text{for } i = 0 \\
  \hat{\mathbf{x}}_{k-1} + (\sqrt{(L+\lambda)\mathbf{P}_{k-1}})_i & \text{for } i = 1, \ldots, L \\
  \hat{\mathbf{x}}_{k-1} - (\sqrt{(L+\lambda)\mathbf{P}_{k-1}})_{i-L} & \text{for } i = L+1, \ldots, 2L
\end{cases}
\end{align}
These points are propagated through the nonlinear state transition function $f(\cdot)$ and the predicted state mean $\hat{\mathbf{x}}_{k|k-1}$ and covariance $\mathbf{P}_{k|k-1}$ are computed by weighted summation.
\begin{align}
\chi_{k|k-1}^{(i)} &= f(\chi_{k-1}^{(i)}, \mathbf{u}_{k-1})
\\
\hat{\mathbf{x}}_{k|k-1} &= \sum_{i=0}^{2L} W_m^{(i)} \chi_{k|k-1}^{(i)}\\
\mathbf{P}_{k|k-1} &= \sum_{i=0}^{2L} W_c^{(i)} \left( \chi_{k|k-1}^{(i)} - \hat{\mathbf{x}}_{k|k-1} \right) \left( \chi_{k|k-1}^{(i)} - \hat{\mathbf{x}}_{k|k-1} \right)^T + \mathbf{Q}
\end{align}
where $\mathbf{Q}$ indicates the process noise covariance matrix, included to account for unknown errors and disturbances in the assumed dynamic model. 

The update or correction step starts by transforming the propagated sigma point  through the measurement function $h(\cdot)$:
\begin{equation}
\gamma_k^{(i)} = h(\chi_{k|k-1}^{(i)})
\end{equation}
The predicted measurement mean $\hat{\mathbf{z}}_k$ and measurement covariance $\mathbf{S}_k$ are computed. Then, the cross-covariance $\mathbf{T}_k$ and Kalman gain $\mathbf{K}_k$ are calculated as

\begin{align}
\hat{\mathbf{z}}_k &= \sum_{i=0}^{2L} W_m^{(i)} \gamma_k^{(i)}\\
\mathbf{S}_k &= \sum_{i=0}^{2L} W_c^{(i)} \left( \gamma_k^{(i)} - \hat{\mathbf{z}}_k \right) \left( \gamma_k^{(i)} - \hat{\mathbf{z}}_k \right)^T + \mathbf{R}
\\
\mathbf{T}_k &= \sum_{i=0}^{2L} W_c^{(i)} \left( \chi_{k|k-1}^{(i)} - \hat{\mathbf{x}}_{k|k-1} \right) \left( \gamma_k^{(i)} - \hat{\mathbf{z}}_k \right)^T\\
\mathbf{K}_k &= \mathbf{T}_k \mathbf{S}_k^{-1}
\end{align}
where $\mathbf{R}$ indicates the measurement noise covariance matrix. Finally, the state estimate $\hat{\mathbf{x}}_k$ and covariance $\mathbf{P}_k$ are updated using the actual measurement $\mathbf{z}_k$:
\begin{align}
\hat{\mathbf{x}}_k = \hat{\mathbf{x}}_{k|k-1} + \mathbf{K}_k (\mathbf{z}_k - \hat{\mathbf{z}}_k)\\
\mathbf{P}_k = \mathbf{P}_{k|k-1} - \mathbf{K}_k \mathbf{S}_k \mathbf{K}_k^T
\end{align}
The state posterior distribution is assumed as Gaussian, and the filter can start a new prediction until the availability of new observations are available for sensor fusion. 

\section{The Measurement Model}
\label{sec:measurement}
The filter processes translational and rotational measurements but propagates the dynamics separately for computational efficiency, requiring minimal hardware footprint. This treats the 12-element state vector as two independent 6-element vectors (translation/velocity and rotation/angular rate), integrated using a 4th-order Runge-Kutta method suitable for onboard use. The filter supports using all visible markers or a limited subset for sensor fusion and can simulate measurement losses to test robustness. 

\subsection{Markers Creation}
For filters to precisely assess the target's attitude and correct the predicted values, precise measurements are necessary. The majority of filters used in space applications rely on image processing from cameras \cite{cassiniRew, pauly, park}. In real-world situations, the image processing program is set up to recognize markers in every picture that is taken. After processing the camera image and determining the marker positions, the software sends this data to the filter.  Given the need for quick image processing, choosing markers is a challenging issue that is impacted by the target's shape, volume, and color.  Target corners are frequently chosen as markers using traditional corner detection algorithms such as the Förstner and Harris-Stephens \cite{chen, wang-1995, NOBLE1988121} methods. It is determined to use the main body's corners as markers and monitor their positions over time in order to solve the ENVISAT relative pose estimation problem. Since each marker's location with respect to the center of mass is known, the marker placements can, therefore, provide useful information. The filter can determine the spacecraft's relative position and velocity as well as recreate its state by following the trajectory of these corners.

\subsection{Camera Transformation and 3D Reconstruction from 2D Pixel Data}
Converting the 2D-pixel coordinates detected by the CNN module into 3D spatial positions is necessary to accurately locate the satellite corners. This back-projection utilizes both intrinsic and extrinsic camera parameters based on a pinhole camera model \cite{corke-2011}.

\subsubsection{Intrinsic Camera Parameters}
The camera's internal characteristics, like its focal length, principal point, and pixel scaling factors, are collectively described by the intrinsic parameters. These are organized into the $3 \times 3$ camera intrinsic matrix, $\mathbf{C}_{int}$, which defines the transformation from 3D coordinates in the camera's reference frame to 2D pixel coordinates on the image plane.
\begin{equation}
\mathbf{C}_{int} = \begin{bmatrix}
f_x & 0   & c_x \\
0   & f_y & c_y \\
0   & 0   & 1
\end{bmatrix}
\end{equation}
Where \( f_x \) and \( f_y \) are the focal lengths in the x and y directions, respectively. \(c_x\) and \(c_y\) represent half the width and height of the image plane, respectively. In the pinhole camera model, they are used to ensure that the \(u\) and \(v\) coordinates are always positive. This transforms the image coordinate system to have a domain of \([0, 2c_x]\) and \([0, 2c_y]\) for \(u\) and \(v\), respectively, rather than \([-c_x, c_x]\) and \([-c_y, c_y]\).

\subsubsection{Extrinsic Camera Parameters}
The extrinsic parameters define the camera's pose within the world coordinate system. These are encapsulated in the extrinsic transformation matrix denoted as $\mathbf{C}_{ext} = [\mathbf{\Gamma}|\mathbf{t}]$. This matrix comprises a $3 \times 3$ rotation matrix, $\mathbf{\Gamma}$, and a $3 \times 1$ translation vector, $\mathbf{t}$. Its function is to convert the 3D world coordinates of an object, such as the satellite, into the camera's own coordinate system according to the following relation:
\begin{equation}
\mathbf{C}_{ext} = \begin{bmatrix}
\mathbf{\Gamma} & \mathbf{t} \\
0 & 1
\end{bmatrix}
\end{equation}
The transformation from world coordinates \( \textbf{X}_{\text{world}} \) to camera coordinates \( \textbf{X}_{\text{camera}} \) is given by:
\begin{equation}
\textbf{X}_{\text{camera}} = \mathbf{\Gamma} \textbf{X}_{\text{world}} + \mathbf{t}
\end{equation}

\subsubsection{Conversion from 2D Pixel Coordinates to 3D World Coordinates}
Reconstructing the 3D positions of the satellite's markers from their 2D image coordinates involves several steps. Initially, the 2D pixel coordinates $(u, v)$ detected by the CNN module are transformed into the camera's coordinate system, denoted as $\textbf{X}_{\text{camera}}$. This first projection is achieved using the intrinsic camera matrix $\mathbf{C}_{int}$ according to the following relation:
\begin{align}
\textbf{X}_{\text{camera}} = \mathbf{C}_{int}^{-1} \begin{bmatrix} u \\ v \\ 1 \end{bmatrix} \cdot \textbf{Z}_{\text{camera}}
\end{align}
This step uses the point's depth and the inverse intrinsic matrix to map 2D pixel coordinates into the 3D camera frame.

\subsubsection{Transformation to World Coordinates}
The computed 3D camera-frame coordinates are then transformed into world coordinates using the extrinsic matrix $\mathbf{C}_{ext}$ according to the relation:
\begin{align}
\textbf{X}_{\text{world}} = \mathbf{\Gamma}^{-1} (\textbf{X}_{\text{camera}} - \mathbf{t})
\end{align}
Through this sequence of transformations, the 2D pixel coordinates detected by the CNN are converted into 3D world coordinates. These resulting 3D points serve as crucial measurements needed for accurate estimation of the satellite's relative position and orientation. Algorithm \ref{algo1} summarizes the whole procedure for the camera transformation as follows.
\begin{algorithm}[H]
\caption{Camera Transformation: 2D Pixel Coordinates $\rightarrow$ 3D World Coordinates}
\label{algo1}
\begin{algorithmic}[1]
\State \textbf{Input:} Pixel coordinates $(u,v)$ from CNN, depth $Z_{\text{camera}}$, 
       intrinsic matrix $\mathbf{C}_{int}$, extrinsic parameters $(\mathbf{\Gamma}, \mathbf{t})$
\State \textbf{Step 1: Back-project pixels into camera frame}
       \[
       \mathbf{X}_{\text{camera}} = \mathbf{C}_{int}^{-1} \begin{bmatrix} u & v & 1 \end{bmatrix}^T \cdot Z_{\text{camera}}
       \]
\State \textbf{Step 2: Transformation into world frame}
       \[
       \mathbf{X}_{\text{world}} = \mathbf{\Gamma}^{-1} \big( \mathbf{X}_{\text{camera}} - \mathbf{t} \big)
       \]
\State \textbf{Output:} 3D world coordinates $\mathbf{X}_{\text{world}}$
\end{algorithmic}
\end{algorithm}

\subsection{Measurement Equations}
\begin{figure}[H]
    \centering
    {\includegraphics[width=0.8\textwidth]{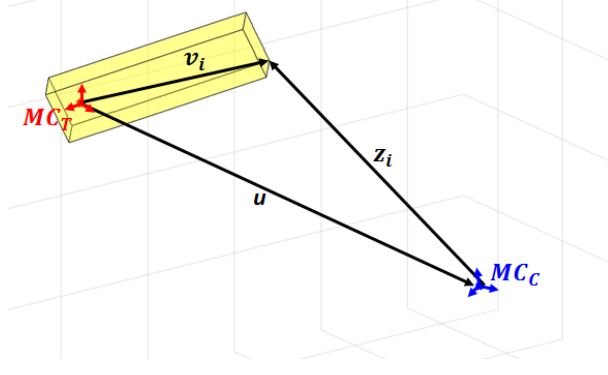}}
    \caption{Arrangement of the vectors for the measurement equation}
    \label{fig:hocam}
\end{figure}

The twelve components that make up the state vector can be separated into four sections. Each section, which is made up of three components, describes one aspect of the relative pose of the target satellite, ENVISAT.
\begin{equation}
\mathbf{x} = (\underbrace{x, y, z}_\text{Pos.}, \underbrace{\dot{x}, \dot{y}, \dot{z}}_\text{Vel.}, \underbrace{p_1, p_2, p_3}_\text{MRP}, \underbrace{w_{x}, w_{y}, w_{z}}_\text{Ang. Vel.})
\end{equation}
It should be noted that the complete simulation uses 18 states for the propagation of absolute states, such as the position and velocity of the chaser's orbit with respect to the ECI frame.  The relative position of the target and chaser's centers of mass, the center of mass's relative velocity, MRPs, and the angular velocities are the four main elements that need to be monitored.  In order to compare the expected measurements with the actual measurements acquired from the camera system during the filter update stage, it is important to assess the marker locations based on the projected state. Let \( \mathbf{u} \) be the position vector of the center of mass of the chaser in relation to the center of mass of ENVISAT. The target's reference frame, which is determined by the available CAD model of ENVISAT, is where the marker position vector \( \mathbf{v}_i \) is first expressed.  After that, this vector is multiplied by the rotation matrix \( \mathbf{\Gamma} \), obtained by the estimated MRP components of the state vector, to create the chaser's reference frame.  Lastly, a straightforward vector difference is used to calculate each marker's position measurement with respect to the chaser as following.
\begin{equation} \label{eq_z}
{\mathbf{z}}_i = \mathbf{\Gamma} \mathbf{v}_i - \mathbf{u}
\end{equation}
where \( {\mathbf{z}}_i \) represents the position measurement of a marker relative to the chaser's center of mass. Thus, this equation indicates that the measurement model couples translation and rotational states.  Fig. \ref{fig:hocam} shows how the vectors are arranged in the measurement equations in a simplified representation of the satellite to help visual understanding \cite{phdthesis}.

\subsection{Markers Association}
Following CNN's detection of the possible marker sites, the estimated states of the filter are used to compare these measurements to the expected locations for each marker as determined by Eq. \eqref{eq_z}. By reducing the overall distance between the predicted and measured places, the association process is executed. To make sure that the anticipated marker sites are as close to the measured ones as possible, the total minimum distance is computed using the Euclidean distance. By building a cost function, \( J \), that minimizes the squared distance between the measured marker positions, \( \mathbf{\Xi}_i \), and the predicted marker locations, \( \mathbf{\tau}_j \), this procedure can be demonstrated mathematically.  To properly associate each predicted marker with the closest measured marker, the cost function \(J\) chooses the pairings \( (i, j) \) that minimize the total squared distance as follows.
\begin{equation}
J = \arg \min_{i,j} \sum_{i,j} (\mathbf{\Xi}_i - \mathbf{\tau}_j)^2
\end{equation}

\subsection{Measurement Noise Covariance Calculation}
To obtain a physically accurate measurement noise covariance matrix $\mathbf{R}$ that reflects real-world sensor uncertainty, a statistical estimation procedure was developed based on repeated CNN detections over a single image, validated for the total image dataset. Rather than assigning arbitrary or constant values to $\mathbf{R}$, we empirically derive it from repeated runs of the corner detection network under simulated variability, resulting in a data-driven characterization of 3D measurement uncertainty. The process begins with a selected image and its corresponding ground truth corner locations and depths, projected from the satellite simulation. CNN detects 2D corner positions over multiple iterations of randomized sensor perturbation. These perturbations include affine jitter, random noise injection, and controlled blur, emulating physical effects such as camera shake, sensor noise, and image degradation. Each set of detected 2D pixel positions is reprojected into 3D coordinates using the camera intrinsic matrix and the associated marker depth values. For each marker $i \in \{A, B, ..., R\}$ (total 18 markers), this process yields a distribution of 3D measurements $\{\mathbf{z}_i^{(k)}\}_{k=1}^{N}$, where $N$ is the number of Monte Carlo CNN runs. The sample covariance of these measurements is then computed as:
\begin{align}
\mathbf{R}_i = \frac{1}{N-1} \sum_{k=1}^{N} (\mathbf{z}_i^{(k)} - \bar{\mathbf{z}}_i)(\mathbf{z}_i^{(k)} - \bar{\mathbf{z}}_i)^T
\end{align}
where $\bar{\mathbf{z}}_i$ is the mean of the 3D positions of marker $i$.

\begin{figure}[H]
    \centering
    \includegraphics[width=\textwidth]{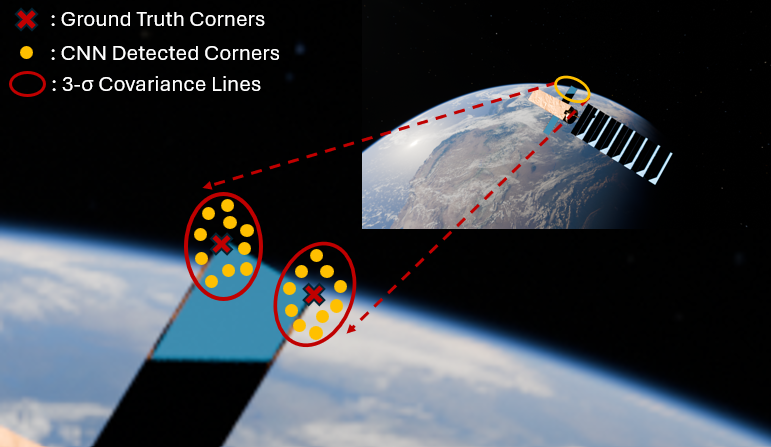}
    \caption{CNN corner detection uncertainty spread under noise}
    \label{fig:marker_spread}
\end{figure}

In order to capture a global characterization of the sensor noise model, all residuals across all markers are stacked, and a global covariance matrix $\mathbf{R}$ is estimated:
\begin{align}
\mathbf{R} = \text{cov} \left( \bigcup_i \left\{ \mathbf{z}_i^{(k)} - \bar{\mathbf{z}}_i \right\}_{k=1}^{N} \right)
\end{align}
The resulting covariance matrix is saved and integrated into the UKF initialization, ensuring the filter's confidence in measurements reflects the observed detection spread. Figure~\ref{fig:marker_spread} visualizes the CNN detected marker spread around the ground truth for a sample image, where ellipses represent 3-sigma contours of the 2D covariance.

\section{Dual Adaptation Strategy for Process and Measurement Noises}
\label{sec:adaptation}
In practice, both the process noise covariance $\mathbf{Q}$ and the measurement noise covariance $\mathbf{R}$ play a central role in determining the stability and accuracy of a Kalman filter. However, conventional approaches typically assume fixed values for these matrices, tuned offline based on heuristics or limited test scenarios \cite{kalmanadap}. Such static assumptions are inadequate in spaceborne relative navigation, where sensor quality and dynamical uncertainties vary significantly with time, lighting conditions, visibility, and eclipse events \cite{ersinhocamadap}. To address this challenge, we develop a dual adaptation strategy that simultaneously tunes $\mathbf{R}$ online through innovation
filtering and adapts $\mathbf{Q}$ during eclipse conditions using a smoothing-inspired formulation. This combined approach ensures both consistency and robustness under realistic and time-varying conditions.

\subsection{Online Adaptive R-Tuning via Innovation Filtering}
Our work introduces an online adaptation strategy for the measurement noise covariance matrix $\mathbf{R}$, which addresses a key limitation in many existing vision-based filtering pipelines: the use of static, overly conservative sensor models. Traditionally, $\mathbf{R}$ is selected offline based on empirical measurements or hand-crafted heuristics, assuming constant sensor quality under time and scene conditions. However, in spaceborne camera tracking, measurement uncertainty is inherently dynamic, affected by marker visibility, occlusions, blur, lighting, and external jitters. Relying on a fixed $\mathbf{R}$ results in either overconfident updates or loss of information gain, both of which degrade filter consistency and estimation quality. In order to overcome this, we introduce an MTF approach, and the theoretical background behind the proposed algorithm is performing an innovation-based adaptive tuning for the measurement noise covariance matrix. Innovation in UKF structure is defined as,
\begin{align}
\mathbf{e}_k=\mathbf{z}_k-\hat{\mathbf{z}}_k,
\end{align}
where ${{\mathbf{e}}}_k$ is the measurement innovation. If there exist mismatches between the process and measurement models, due to the unaccounted external disturbances during the gathering of measurements, the Kalman gain changes with varying innovation covariance. For an optimally performing Kalman filter operating in steady-state, the presence of external disturbances affecting the measurements coming from the camera is indicated when the trace of the actual error covariance surpasses a theoretical value. This condition, requiring the activation of adaptive filtering, is given by:
\begin{align}
\label{eq:adaptive_trigger} 
\text{tr}({\mathbf{e}}_k{\mathbf{e}}_k^T) \geq \text{tr}(\mathbf{S}_{k-1}+\mathbf{MTF}_k+\mathbf{R}),
\end{align}

where $\text{tr}(\cdot)$ denotes the matrix trace. Specifically, monitoring the real innovation covariance ($\mathbf{e}_k\mathbf{e}_k^T$) at the current time $k$ allows for the detection of instantaneous variations. So, from this point, at each update step, we compute:
\begin{align}
\mathbf{MTF}_k = \mathbf{e}_{k} \mathbf{e}_{k}^T - \mathbf{S}_{k-1} - \mathbf{R}
\end{align}
Instead of using the full $\mathbf{MTF}_k$ matrix, only its diagonal elements are considered for the adaptive covariance update. This simplification arises from the assumption that external disturbances affecting different markers are uncorrelated. Furthermore, since covariance must be non-negative for physical significance, any negative diagonal values computed in $\mathbf{MTF}_k$ are subsequently set to zero. This positivity constraint is enforced element-wise, represented as:
\begin{align}
\mathbf{MTF}_{k} \leftarrow \max(0, \mathbf{MTF}_k)
\end{align}
where the $\max(\cdot)$ operation ensures all elements are non-negative. This matrix captures underestimation in $\mathbf{R}$ during those camera frame acquisitions with higher noise than usual. Therefore, the measurement covariance now becomes as follows:
\begin{align}
\mathbf{S}_{k} = \mathbf{S}_{k-1} + \mathbf{R} + \mathbf{MTF}_{k}
\end{align}

Therefore, instead of using batch residual statistics to adjust $\mathbf{R}$ over long windows, our formulation is fully online and adaptive, enabling efficient tuning in real-time, even under partial marker loss, or sudden lighting changes.

\subsection{Online Adaptive Q-Tuning with Smoothing Approach}
In our previous study \cite{candan}, we addressed the problem of missing observations due to eclipses by manually adjusting the process noise covariance matrix $\mathbf{Q}$ during known periods of the eclipse. This adjustment was based on prior knowledge of the eclipse interval, during which measurement updates were entirely disabled. To compensate for the increased uncertainty caused by the absence of visual measurements, the values in $\mathbf{Q}$ were heuristically inflated, allowing the UKF to propagate broader state uncertainty. While this approach proved effective in maintaining robust estimates under eclipse conditions, it relied heavily on manual tuning and a priori knowledge of eclipse duration. In contrast, the method introduced in this work leverages the UKF's native sigma point structure to compute a forward-looking cross-covariance matrix $\mathbf{D}_k$, enabling adaptive $\mathbf{Q}$ updates via a closed-form expression inspired by RTS smoothing, which the original algorithm, Algorithm \ref{algo2} is shown as follows \cite{Sarkka2013}. The key idea behind smoothing is to refine state estimates by taking into account not only past and current measurements but also future information. In contrast to the standard filtering approach, which propagates estimates forward in time using local process and measurement models, smoothing revisits earlier estimates after the entire trajectory is available. This allows the algorithm to reduce estimation error by 
averaging out the effect of process and measurement noise across multiple 
time steps.

\begin{algorithm}[H]
\caption{Unscented RTS Smoother}
\label{algo2}
\begin{algorithmic}[1]
\Require $f,\mathbf{Q},h,\mathbf{R}$, UT params $(\alpha,\beta,\kappa)$, prior $(\textbf{x}_0,\mathbf{P}_0)$
\For{$k=0$ to $T-1$}
  \State Build UT weights and sigma points $\chi_k^{(i)}$
  \State ${\chi}_{k|k-1}^{(i)} \gets f(\chi_{k-1}^{(i)})$
  \State $\hat{\textbf{x}}_{k+1} \gets \sum W_m^{(i)} {\chi}_{k|k-1}^{(i)}$
  \State $\textbf{P}_{k|k-1} \gets \sum W_c^{(i)}({\chi}_{k|k-1}^{(i)}-\hat{\textbf{x}}_{k|k-1})(\cdot)^\top + \mathbf{Q}$
  \State ${\gamma}_{k}^{(i)} \gets h({\chi}_{k|k-1}^{(i)})$, $\hat{\textbf{z}}_{k} \gets \sum W_m^{(i)}{\gamma}_{k}^{(i)}$
  \State $\textbf{S}_{k} \gets \sum W_c^{(i)}({\gamma}_{k}^{(i)}-\hat{\textbf{z}}_{k})(\cdot)^T + \mathbf{R}$
  \State $\textbf{T}_{k} \gets \sum W_c^{(i)}({\chi}_{k|k-1}^{(i)}-\hat{\textbf{x}}_{k|k-1})(\gamma_{k}^{(i)}-\hat{\textbf{z}}_{k})^T$
  \State $\textbf{K}_{k} \gets \textbf{T}_{k}\textbf{S}_{k}^{-1}$;\quad
         $\textbf{x}_{k} \gets \textbf{x}_{k|k-1}+\textbf{K}_{k}(\textbf{z}_{k}-\hat{\textbf{z}}_{k})$
  \State $\textbf{P}_{k+1} \gets \textbf{P}_{k|k-1}-\textbf{K}_{k}\textbf{S}_{k}\textbf{K}_{k}^T$
  \State $\textbf{D}_{k} \gets \sum W_c^{(i)}(\chi_{k-1}^{(i)}-\hat{\textbf{x}}_{k-1})(\hat{\chi}_{k|k-1}^{(i)}-\hat{\textbf{x}}_{k})^T$
\EndFor
\State $\textbf{x}_T^s \gets \textbf{x}_T$, $\textbf{P}_T^s \gets \textbf{P}_T$
\For{$k=T-1$ down to $0$}
  \State $\textbf{G}_k \gets \textbf{D}_{k+1}\textbf{P}_{k+1|k}^{-1}$
  \State $\textbf{x}_k^s \gets \textbf{x}_k + \textbf{G}_k(\textbf{x}_{k+1}^s-\hat{\textbf{x}}_{k+1})$
  \State $\textbf{P}_k^s \gets \textbf{P}_k + \textbf{G}_k(\textbf{P}_{k+1}^s-\textbf{P}_{k+1|k})\textbf{G}_k^T$
\EndFor
\State \Return $\{\textbf{x}_k^s,\textbf{P}_k^s\}_{k=0}^T$
\end{algorithmic}
\end{algorithm}

Unlike using the whole UKF-RTS algorithm in our work, which will require a back propagation to improve the accuracy of estimates at previous time steps, we will make use of the cross-covariance between sigma points propagated forward in time. In order to dynamically tune $\mathbf{Q}$ at each filter step, $\mathbf{D}$ is computed at each step as following.
\begin{align}
\mathbf{D}_{k} = W_c \left({\chi}_{k-1} - \hat{\mathbf{x}}_{k-1} \right) \left( {\chi}_{k|k-1} - \hat{\mathbf{x}}_{k} \right)^T
\end{align}
where ${\chi}_{k-1}$ and ${\chi}_{k|k-1}$ denote sigma points at times $k$ and $k-1$, respectively, and $W_c$ are the unscented transform (UT) weights. This matrix captures how perturbations at time $k$ influence the state evolution. During periods of no measurements, $\mathbf{D}$ naturally reflects increased sensitivity and model uncertainty. Instead of inflating $\mathbf{Q}$ manually or based on hard-coded time windows, we adapt $\mathbf{Q}$ online through the additive correction in case of no measurement, no update scenarios, as follows.

\begin{align}
\mathbf{Q}_{\text{adaptive}, k} = \mathbf{D}_k  (\mathbf{P}_{k|k-1} - \mathbf{P}_{k-1})   \mathbf{D}_k^T\\
\mathbf{P}_{k} = \mathbf{P}_{k|k-1}+\mathbf{Q}_{\text{adaptive}, k}
\end{align}

The analogies with the UKF smoother presented in \cite{Sarkka2013} are evident. The proposed method assumes that the system dynamics are sufficiently smooth such that the UT remains locally accurate. Under no severe nonlinearity, the sigma points generated by the UKF accurately capture the shape and spread of the propagated probability density. As a result, the cross-covariance term \( \mathbf{D}_k \) formed between pre- and post-propagation sigma points correctly captures the dominant directions of uncertainty growth. The difference \( \mathbf{P}_{k|k-1} - \mathbf{P}_{k-1} \) measures the increase in uncertainty due to process dynamics and/or model mismatch, allowing the adaptive update. Moreover, that difference captures how much the uncertainty has grown during the propagation phase, while \( \mathbf{D}_k \) acts as a projection operator that aligns this growth with the axes along which the sigma points actually diverged. The resulting update matrix thus injects noise in the directions most sensitive to model error, leading to improved filter consistency and resilience to unmodeled dynamics. This non-iterative, real-time update requires no batch statistics or tuning windows and automatically responds to dynamic model mismatches or visibility loss (e.g., eclipse scenarios), providing robust performance with minimal computations. Compared to previous methods, our approach scales efficiently to high-dimensional state spaces and is well-suited for embedded onboard applications in vision-based space navigation.  This formulation ensures that increased uncertainty due to the no-markers and no-measurement case is captured efficiently by the filter. Moreover, the update term functions as a directionally aware, low-rank covariance inflator. This proposed adaptive technique allows for real-time noise adaptation while keeping the computational cost of the filter low. Indeed, the cross-correlation matrix makes use of already available states, as the sigma point must be propagated independently from the adaptation technique.

\section{Numerical Results}
\label{sec:results}
To initialize the estimation framework, uncertainty is modeled by adding zero-mean Gaussian noise to the initial state vector. The standard deviations applied to each component reflect assumed initial confidence levels, as specified in Table~\ref{tab:initial_conditions}. A normal distribution with a standard deviation of $1.0 m$ for position components, $1.0 \times 10^{-1} m/s$  for velocity, $1.0 \times 10^{-2} rad$ for MRPs, and $1.0 \times 10^{-3} rad/s$ for angular velocity. The estimation framework is tested with the Monte Carlo runs. For each Monte Carlo analysis, the red dashed lines represent 3-$\sigma$ bounds from the approach with no adaptation, and the solid blue lines indicate the predicted 3-$\sigma$ bounds from the corresponding adaptive filter. Monte Carlo simulation errors are shown in gray, where darker gray represents the adaptation Monte Carlo errors, while lighter gray represents no adaptation Monte Carlo errors. The solid black line represents the adaptive mean error, while the dashed black line represents the no-adaptation mean error.

\begin{table}{
    \caption{Initial conditions for Monte Carlo simulations}
    \label{tab:initial_conditions}
    \centering
    \begin{tabularx}{\textwidth}{@{}l >{\raggedleft\arraybackslash}X | l >{\raggedleft\arraybackslash}X@{}}
        \hline
        \multicolumn{2}{c|}{\textbf{Translational Dynamics}} & \multicolumn{2}{c}{\textbf{Rotational Dynamics}} \\
        \hline
        $x$ (m)           & -0.002       & $\phi$ (roll,rad)           & 1.66  \\
        $y$ (m)           & -31.17       & $\theta$ (pitch,rad)         & 2.27  \\
        $z$ (m)           & 0            & $\psi$ (yaw,rad)           & -0.38 \\
        $\dot{x}$ (m/s)   & -3.5e-6      & $\omega_{x}$ (rad/s) & 0.02  \\
        $\dot{y}$ (m/s)   & -2.0e-6      & $\omega_{y}$ (rad/s) & 0.02  \\
        $\dot{z}$ (m/s)   & 0            & $\omega_{z}$ (rad/s) & 0.04  \\
        \hline
    \end{tabularx}}
\end{table}

\begin{figure}[H]
    \centering    
    \includegraphics[width=\textwidth]{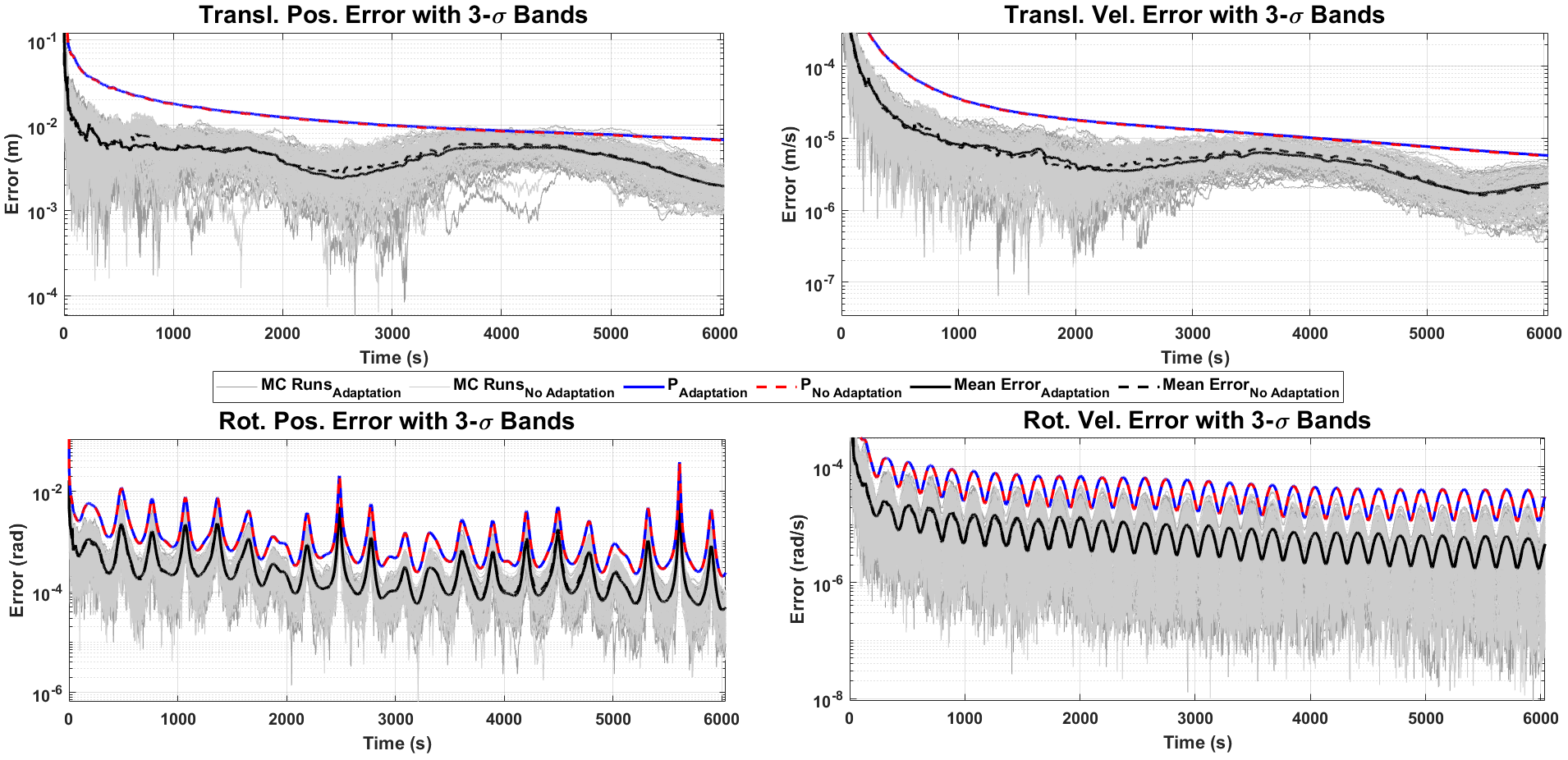}
    \caption{RMSE comparison for MTF and no adaptation cases for the no-eclipse scenario}
    \label{fig:RMSE_NoEclipse_MTF}
\end{figure}

\begin{figure}[H]
    \centering
    \includegraphics[width=\textwidth]{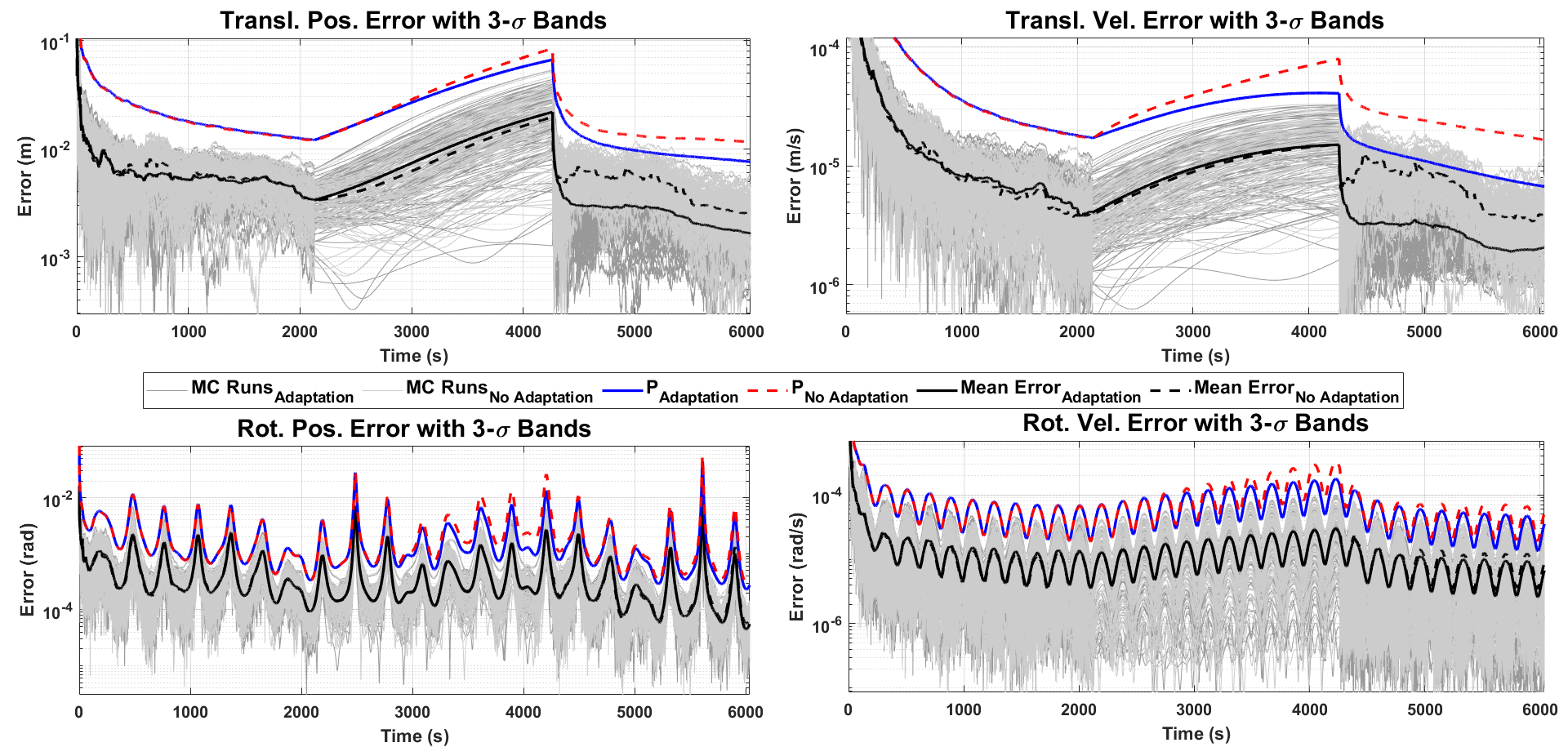}
    \caption{RMSE comparison for $\textbf{MTF}+\mathbf{Q}_{adaptation}$ and no adaptation cases for the eclipse scenario}
    \label{fig:RMSE_Qadapt}
\end{figure}

The adaptive \textbf{R}-tuning (using \textbf{MTF}) enhances the estimation quality of the UKF, allowing it to trust high confidence measurements while guarding against noisy or spurious detections. As demonstrated in the Monte Carlo analysis during the no-eclipse scenario, the adaptive filter maintains tighter error bounds and more accurate pose estimation than the no-adaptation case for both scenarios with time-varying measurement noise. Fig. \ref{fig:RMSE_NoEclipse_MTF} shows the root mean squared errors (RMSE) for translational and rotational states as follows, where the log scale is used for ease of understanding.
The RMSE we use in this is study is defined as,
\begin{equation}
\mathrm{RMSE}(k) \;=\; 
\sqrt{\frac{1}{N_m} \sum_{j=1}^{N_m} 
\big\| \mathbf{s}_{k,j} - \hat{\mathbf{s}}_{k,j} \big\|^2 }
\end{equation}
where $\mathbf{s}_{k,j}$ and $\hat{\mathbf{s}}_{k,j}$ denote the ground-truth 
and estimated state vectors, respectively, at time step $k$ for the $j$-th Monte Carlo run, and $N_m$ is the total number of runs. For translational errors, the Euclidean norm is taken over the position components, while for rotational errors it is taken over the attitude or angular velocity components.This adaptive \textbf{R}-tuning significantly enhances the consistency of the UKF, allowing it to trust high-confidence measurements while guarding against noisy or spurious detections. As demonstrated in the Monte Carlo analysis during the no-eclipse scenario, the adaptive filter maintains tighter error bounds and more accurate pose estimation than the no-adaptation case for both scenarios with time-varying measurement noise. Fig. \ref{fig:RMSE_Qadapt} shows the RMSE for translational and rotational states as follows. For the ENVISAT satellite, the eclipse duration was calculated to be 35.57 minutes. The most notable improvement is observed during and after the eclipse interval, starting at $35.5^{th}$ minute, where the adaptive filter maintains tighter 3-$\sigma$ bounds that closely track the actual error evolution. On the other hand, the non-adaptive case shows a growing divergence between predicted covariance and actual error. Particularly in each channel, the $\mathbf{Q}_{\text{adaptive}} +\textbf{MTF}$ method significantly reduces over-conservativeness in $\mathbf{P}$ while simultaneously preventing underestimation of state uncertainty. These results confirm the effectiveness of our data-driven, forward-smoothed $\mathbf{Q}_{\text{adaptive}} +\textbf{MTF}$ strategy, which enables the UKF to retain robustness and confidence during long measurement outages.

To comprehensively evaluate the impact of the proposed dual adaptation strategy, Table~\ref{tab:rmse_compact} presents a detailed comparison of RMSE values for all 12 states under both Eclipse (E) and No Eclipse (NE) conditions. The results are computed across multiple Monte Carlo simulations and are reported for different filter configurations. UKF with no adaptation, filters employing either $\mathbf{R}$ or $\mathbf{Q}$ adaptation alone, and the proposed fully adaptive filter with simultaneous $\mathbf{Q}_{\text{adaptive}} +\textbf{MTF}$ tuning.

\begin{tableorg}
\caption{RMSE values under Eclipse (E) and No Eclipse (NE) conditions for different methods}
\label{tab:rmse_compact}
\begin{center}
\begin{threeparttable}

\begin{adjustbox}{max width=\textwidth}
\begin{tabular}{c l l @{\qquad} c l l}
\toprule
\textbf{State} & \textbf{Method} & \textbf{RMSE (E / NE)} &
\textbf{State} & \textbf{Method} & \textbf{RMSE (E / NE)} \\
\midrule

\multirow{4}{*}{$x$}
& MTF + $\mathbf{Q}_{\mathrm{adaptive}}$ & $\mathbf{0.00842} / -$~m
& \multirow{4}{*}{$p_1$}
& MTF + $\mathbf{Q}_{\mathrm{adaptive}}$ & $\mathbf{8.571\!\times\!10^{-4}} / -$~rad \\
\cmidrule(lr){2-3}\cmidrule(lr){5-6}
& MTF & $\mathbf{0.00955} / \mathbf{0.00623}$~m
& & MTF & $\mathbf{8.654\!\times\!10^{-4}} / \mathbf{6.556\!\times\!10^{-4}}$~rad \\
\cmidrule(lr){2-3}\cmidrule(lr){5-6}
& $\mathbf{Q}_{\mathrm{adaptive}}$ & $0.01034 / -$~m
& & $\mathbf{Q}_{\mathrm{adaptive}}$ & $0.001306 / -$~rad \\
\cmidrule(lr){2-3}\cmidrule(lr){5-6}
& No Adaptation & $0.01393 / 0.00888$~m
& & No Adaptation & $0.001531 / 0.001393$~rad \\
\midrule

\multirow{4}{*}{$y$}
& MTF + $\mathbf{Q}_{\mathrm{adaptive}}$ & $\mathbf{0.00922} / -$~m
& \multirow{4}{*}{$p_2$}
& MTF + $\mathbf{Q}_{\mathrm{adaptive}}$ & $\mathbf{9.823\!\times\!10^{-4}} / -$~rad \\
\cmidrule(lr){2-3}\cmidrule(lr){5-6}
& MTF & $\mathbf{0.009427} / \mathbf{0.00467}$~m
& & MTF & $\mathbf{9.756\!\times\!10^{-4}} / \mathbf{7.519\!\times\!10^{-4}}$~rad \\
\cmidrule(lr){2-3}\cmidrule(lr){5-6}
& $\mathbf{Q}_{\mathrm{adaptive}}$ & $0.01102 / -$~m
& & $\mathbf{Q}_{\mathrm{adaptive}}$ & $0.001393 / -$~rad \\
\cmidrule(lr){2-3}\cmidrule(lr){5-6}
& No Adaptation & $0.01142 / 0.00488$~m
& & No Adaptation & $0.001532 / 0.001404$~rad \\
\midrule

\multirow{4}{*}{$z$}
& MTF + $\mathbf{Q}_{\mathrm{adaptive}}$ & $\mathbf{0.006722} / -$~m
& \multirow{4}{*}{$p_3$}
& MTF + $\mathbf{Q}_{\mathrm{adaptive}}$ & $\mathbf{9.815\!\times\!10^{-4}} / -$~rad \\
\cmidrule(lr){2-3}\cmidrule(lr){5-6}
& MTF & $\mathbf{0.006533} / \mathbf{0.006136}$~m
& & MTF & $\mathbf{9.699\!\times\!10^{-4}} / \mathbf{7.432\!\times\!10^{-4}}$~rad \\
\cmidrule(lr){2-3}\cmidrule(lr){5-6}
& $\mathbf{Q}_{\mathrm{adaptive}}$ & $0.01980 / -$~m
& & $\mathbf{Q}_{\mathrm{adaptive}}$ & $0.001418 / -$~rad \\
\cmidrule(lr){2-3}\cmidrule(lr){5-6}
& No Adaptation & $0.01989 / 0.02247$~m
& & No Adaptation & $0.001606 / 0.001396$~rad \\
\midrule

\multirow{4}{*}{$\dot{x}$}
& MTF + $\mathbf{Q}_{\mathrm{adaptive}}$ & $\mathbf{1.725\!\times\!10^{-4}} / -$~m/s
& \multirow{4}{*}{$\omega_x$}
& MTF + $\mathbf{Q}_{\mathrm{adaptive}}$ & $\mathbf{3.123\!\times\!10^{-5}} / -$~rad/s \\
\cmidrule(lr){2-3}\cmidrule(lr){5-6}
& MTF & $\mathbf{1.740\!\times\!10^{-4}} / \mathbf{1.719\!\times\!10^{-4}}$~m/s
& & MTF & $\mathbf{3.225\!\times\!10^{-5}} / \mathbf{2.634\!\times\!10^{-5}}$~rad/s \\
\cmidrule(lr){2-3}\cmidrule(lr){5-6}
& $\mathbf{Q}_{\mathrm{adaptive}}$ & $1.545\!\times\!10^{-4} / -$~m/s
& & $\mathbf{Q}_{\mathrm{adaptive}}$ & $3.780\!\times\!10^{-5} / -$~rad/s \\
\cmidrule(lr){2-3}\cmidrule(lr){5-6}
& No Adaptation & $1.601\!\times\!10^{-4} / 1.553\!\times\!10^{-4}$~m/s
& & No Adaptation & $4.383\!\times\!10^{-5} / 3.048\!\times\!10^{-5}$~rad/s \\
\midrule

\multirow{4}{*}{$\dot{y}$}
& MTF + $\mathbf{Q}_{\mathrm{adaptive}}$ & $\mathbf{8.331\!\times\!10^{-5}} / -$~m/s
& \multirow{4}{*}{$\omega_y$}
& MTF + $\mathbf{Q}_{\mathrm{adaptive}}$ & $\mathbf{4.630\!\times\!10^{-5}} / -$~rad/s \\
\cmidrule(lr){2-3}\cmidrule(lr){5-6}
& MTF & $\mathbf{8.482\!\times\!10^{-5}} / \mathbf{7.932\!\times\!10^{-5}}$~m/s
& & MTF & $\mathbf{4.773\!\times\!10^{-5}} / \mathbf{3.804\!\times\!10^{-5}}$~rad/s \\
\cmidrule(lr){2-3}\cmidrule(lr){5-6}
& $\mathbf{Q}_{\mathrm{adaptive}}$ & $7.641\!\times\!10^{-5} / -$~m/s
& & $\mathbf{Q}_{\mathrm{adaptive}}$ & $5.956\!\times\!10^{-5} / -$~rad/s \\
\cmidrule(lr){2-3}\cmidrule(lr){5-6}
& No Adaptation & $8.138\!\times\!10^{-5} / 7.330\!\times\!10^{-5}$~m/s
& & No Adaptation & $6.634\!\times\!10^{-5} / 4.878\!\times\!10^{-5}$~rad/s \\
\midrule

\multirow{4}{*}{$\dot{z}$}
& MTF + $\mathbf{Q}_{\mathrm{adaptive}}$ & $\mathbf{1.692\!\times\!10^{-4}} / -$~m/s
& \multirow{4}{*}{$\omega_z$}
& MTF + $\mathbf{Q}_{\mathrm{adaptive}}$ & $\mathbf{2.920\!\times\!10^{-5}} / -$~rad/s \\
\cmidrule(lr){2-3}\cmidrule(lr){5-6}
& MTF & $\mathbf{1.694\!\times\!10^{-4}} / \mathbf{1.688\!\times\!10^{-4}}$~m/s
& & MTF & $\mathbf{3.025\!\times\!10^{-5}} / \mathbf{2.453\!\times\!10^{-5}}$~rad/s \\
\cmidrule(lr){2-3}\cmidrule(lr){5-6}
& $\mathbf{Q}_{\mathrm{adaptive}}$ & $1.884\!\times\!10^{-4} / -$~m/s
& & $\mathbf{Q}_{\mathrm{adaptive}}$ & $3.560\!\times\!10^{-5} / -$~rad/s \\
\cmidrule(lr){2-3}\cmidrule(lr){5-6}
& No Adaptation & $1.882\!\times\!10^{-4} / 1.902\!\times\!10^{-4}$~m/s
& & No Adaptation & $4.088\!\times\!10^{-5} / 2.884\!\times\!10^{-5}$~rad/s \\
\bottomrule
\end{tabular}
\end{adjustbox}


\end{threeparttable}
\end{center}
\end{tableorg}

The results presented in Table \ref{tab:rmse_compact} show the superior performance of the proposed dual adaptive UKF framework in both eclipse and nominal lighting conditions. This comparison reveals the effectiveness of each adaptation strategy in reducing estimation error across diverse operational phases. Notably, the full \textbf{MTF}+$\mathbf{Q}_{\text{adaptive}}$ framework achieves the lowest RMSE in nearly every state, with the bold entries in the table indicating the adaptive strategy values for each state dimension, typically corresponding to the best performance. The table highlights how joint adaptation improves robustness and estimation precision beyond what isolated adaptation strategies can achieve. These enhancements come without sacrificing stability, making the filter robust and reliable for real-world spaceborne navigation scenarios, especially under intermittent visibility, external disturbances, or camera degradation.

\begin{figure}[H]
    \centering    
    \includegraphics[width=\textwidth]{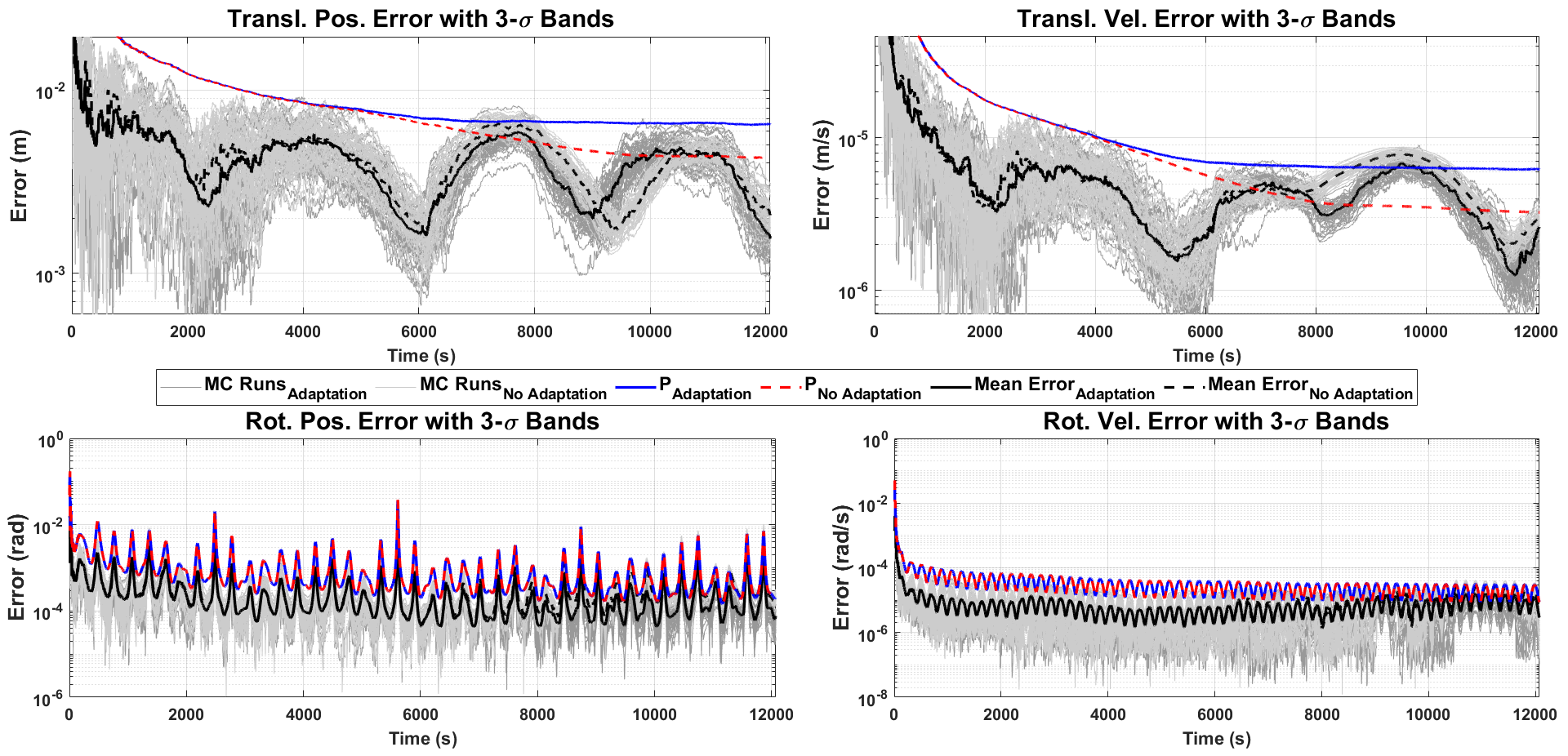}
    \caption{RMSE comparison for MTF and no adaptation cases for the no-eclipse scenario at 0.1~Hz measurement update rate}
    \label{fig:RMSE_NoEclipse_uzun}
\end{figure}

\begin{figure}[H]
    \centering    
    \includegraphics[width=\textwidth]{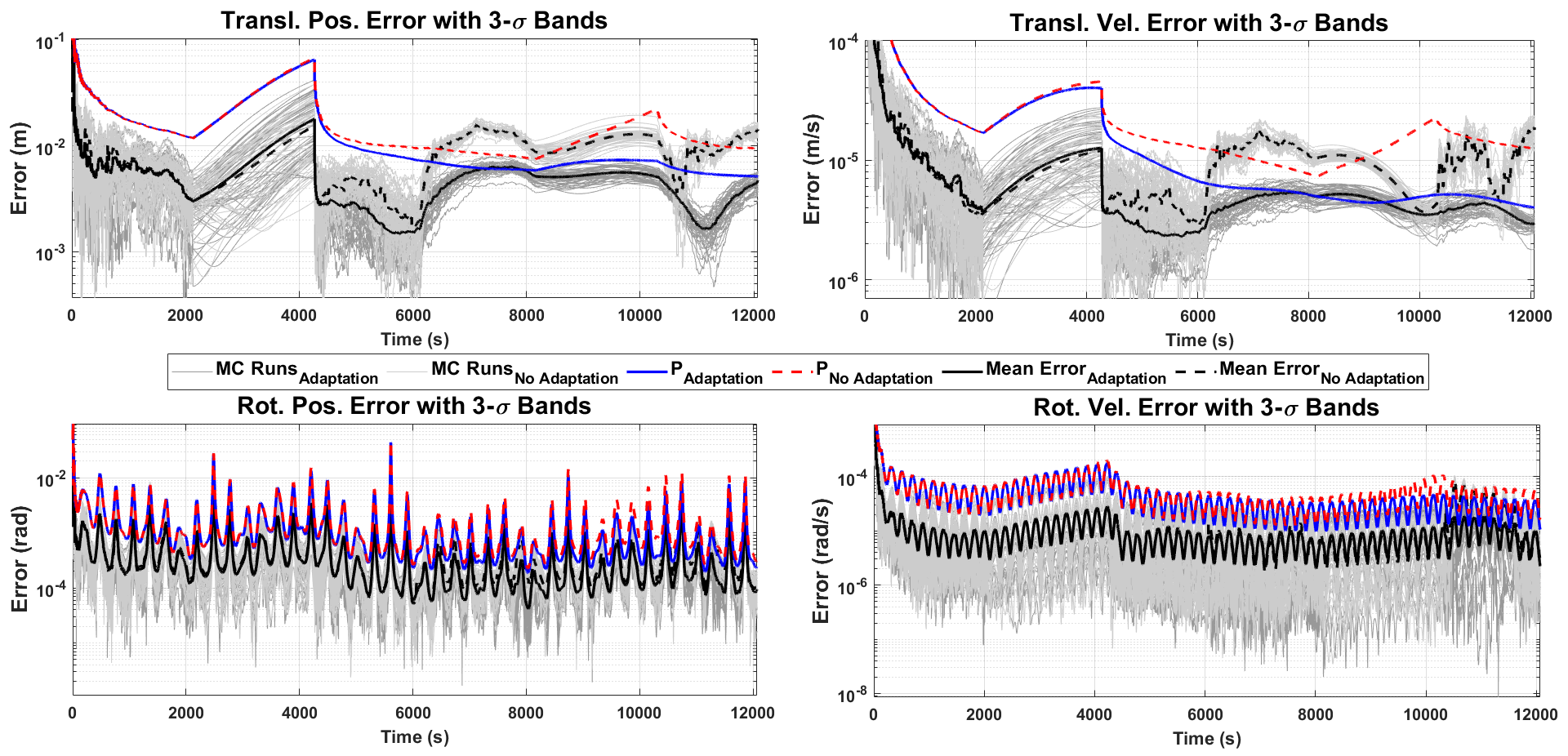}
\caption{RMSE comparison for MTF + $\textbf{Q}_{adaptation}$ and no adaptation cases for the eclipse scenario at 0.1~Hz measurement update rate}
    \label{fig:RMSE_MTFD_uzun}
\end{figure}

In addition to the compact RMSE summary in Table~\ref{tab:rmse_compact}, we also performed Monte Carlo simulations with a reduced measurement acquisition frequency of 0.1~Hz, while maintaining a filter propagation rate of 1~Hz. This scenario represents a more realistic condition in which sensor updates arrive intermittently or with latency. The results, shown in Fig.~\ref{fig:RMSE_NoEclipse_uzun} without any eclipse and Fig.~\ref{fig:RMSE_MTFD_uzun}, confirm that the dual adaptive framework remains robust even under delayed measurement updates. Moreover, simulations were extended to cover two full orbital periods, allowing evaluation across multiple eclipse phases as seen on Fig.~\ref{fig:RMSE_MTFD_uzun}. Notably, the performance gap between the adaptive and non-adaptive filters becomes especially pronounced after the first eclipse, where the adaptive filter successfully recovers with bounded errors, while the non-adaptive filter accumulates significant divergence. These findings emphasize that the proposed dual adaptation not only improves accuracy under nominal conditions, but also sustains robustness during long-duration operations with measurement delays and repeated eclipse events.
Note that,for the no-adaptation case on Fig.~\ref{fig:RMSE_NoEclipse_uzun}, the filter becomes overconfident, the $3\sigma$ bounds fall below the actual error trajectories especially during the second orbit period. This mismatch indicates that the covariance is underestimated, leading to inconsistency and unreliable state uncertainty quantification.

\begin{figure}[H]
    \centering    
    \includegraphics[width=\textwidth]{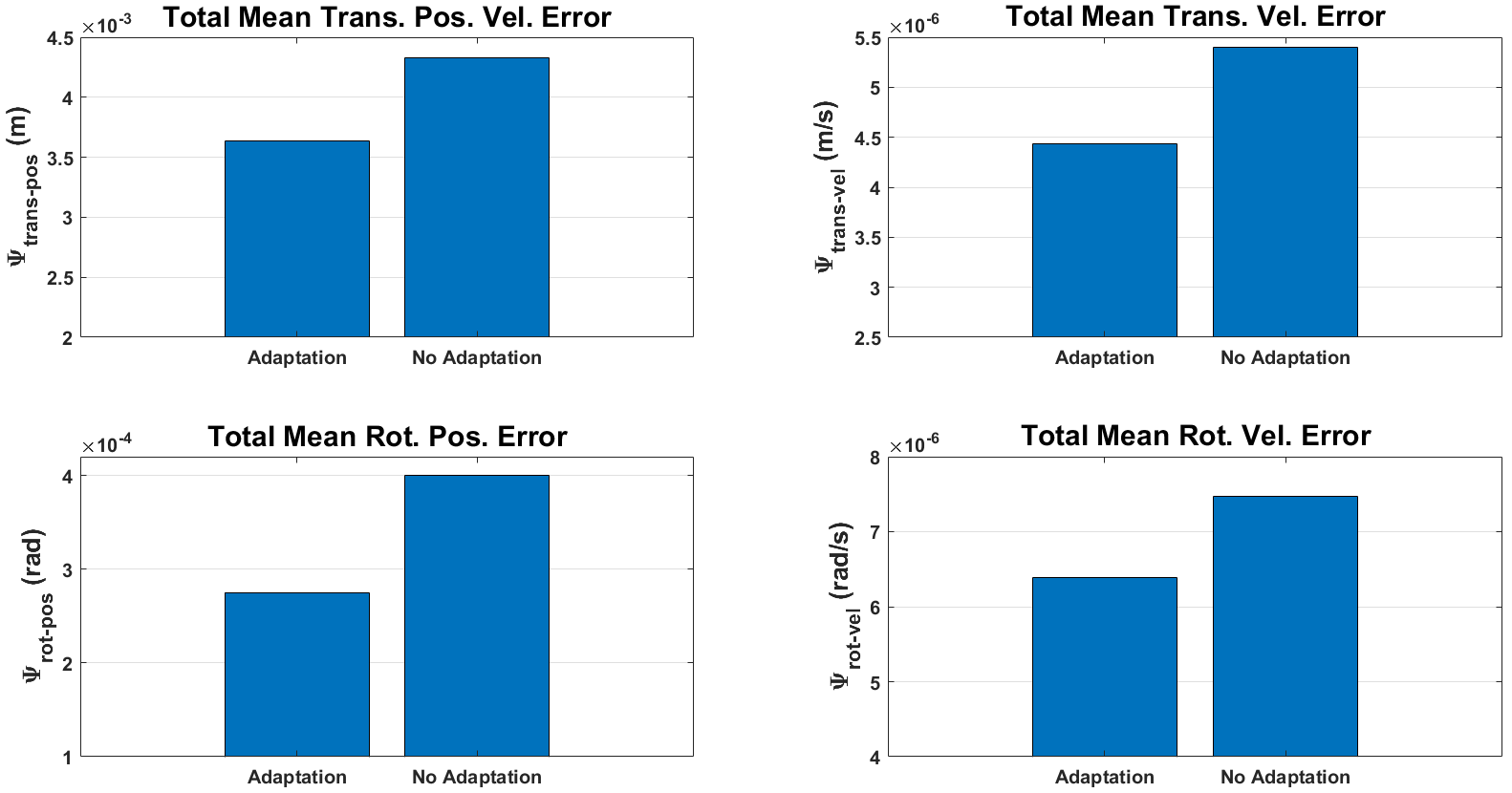}
\caption{ Bar chart RMSE comparison summary for MTF and no adaptation cases for the no-eclipse scenario at
0.1 Hz measurement update rate}
    \label{fig:uzun_NE_bar}
\end{figure}

\begin{figure}[H]
    \centering    
    \includegraphics[width=\textwidth]{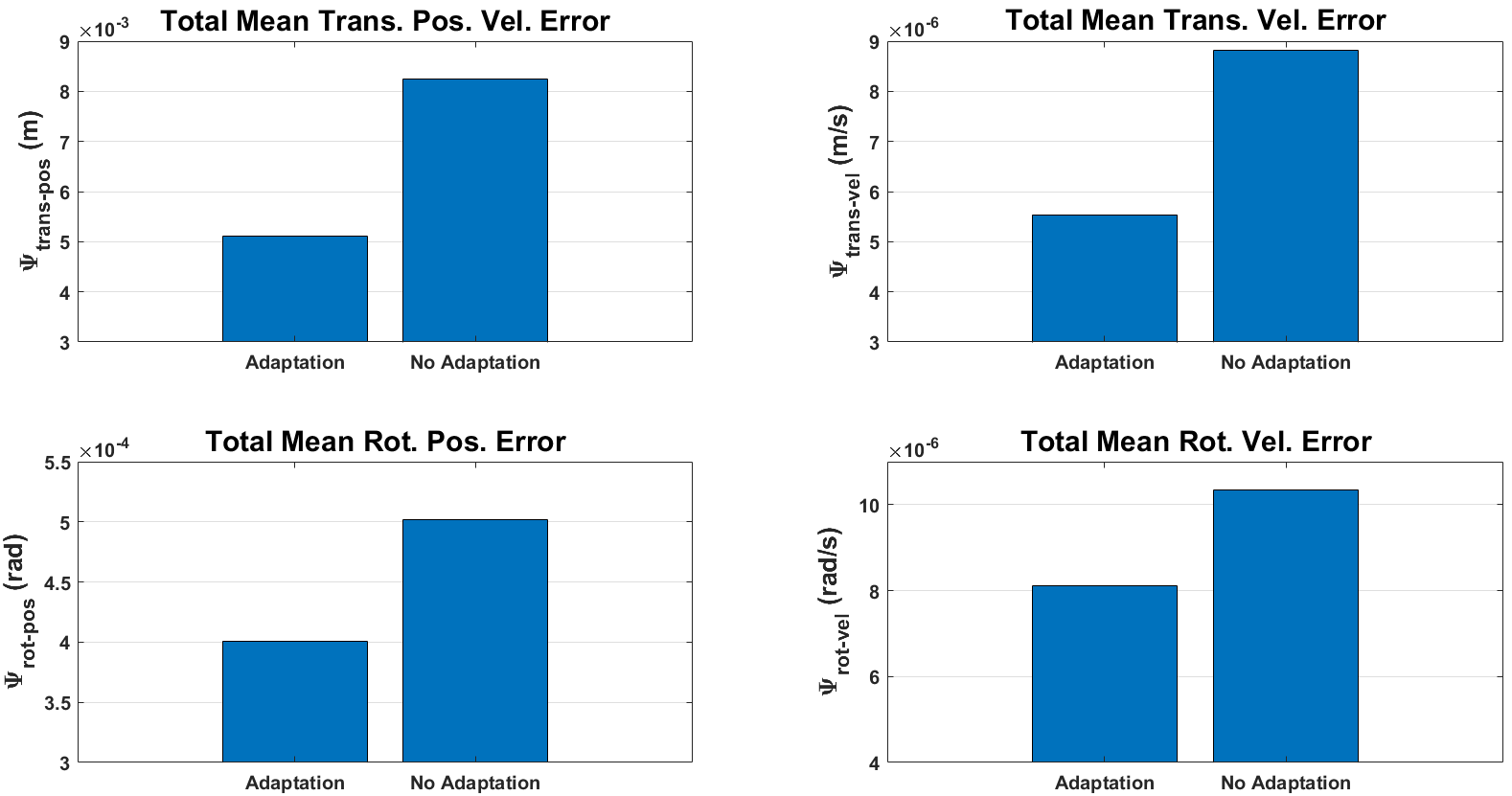}
\caption{Bar chart RMSE comparison summary for MTF + $\textbf{Q}_{adaptation}$ and no adaptation cases for the eclipse scenario at 0.1~Hz measurement update rate}
    \label{fig:uzun_E_bar}
\end{figure}

In order to clearly demonstrate the superior accuracy of the adaptations, we also want to report the total mean translational and rotational errors, both in position and velocity, with and without adaptation for no-eclipse and eclipse cases, respectively. The total mean error, denoted by $\Psi$, is defined as,
\begin{equation}
\Psi = \frac{1}{N_m \, N_t} \sum_{j=1}^{N_m} \sum_{k=1}^{N_t} 
\left\| \mathbf{s}_{k,j} - \hat{\mathbf{s}}_{k,j} \right\|
\end{equation}
where $\mathbf{s}_{k,j}$ and $\hat{\mathbf{s}}_{k,j}$ denote the ground-truth 
and estimated state vectors, respectively, at time step $k$ for the $j$-th Monte Carlo run similar to the ones defined for RMSE formulation. Here, $N_m$ is the number of Monte Carlo runs and $N_t$ is the number of time steps. This scalar quantity $\Psi$ provides an aggregate 
measure of the average error across the entire simulation horizon and across all Monte Carlo realizations. In all cases, the adaptive filter achieves substantially lower errors. These results confirm that adaptation enhances accuracy, stability, and robustness in every scenario. For clarity, we also distinguish between different state components by introducing dedicated symbols. Specifically, the total mean translational 
position error is denoted as $\Psi_{\text{trans-pos}}$, the translational velocity error as $\Psi_{\text{trans-vel}}$, the rotational position error as 
$\Psi_{\text{rot-pos}}$, and the rotational velocity error as $\Psi_{\text{rot-vel}}$. These metrics are directly reported in Figs.~\ref{fig:uzun_NE_bar} and \ref{fig:uzun_E_bar}, providing a compact yet 
comprehensive comparison between adaptive and non-adaptive filters under both conditions.

\section{Measurement Model Improvement}
\label{sec:meas_improvement}
Light Detection and Ranging (LiDAR) has become a central sensing strategy for spacecraft relative navigation, particularly in scenarios involving non-cooperative targets such as orbital debris or uncooperative satellites. Unlike monocular cameras, which provide only bearing information, LiDAR directly measures range and can generate dense three-dimensional point clouds of an object's surface. This capability
enables robust pose estimation even under variable lighting or partial occlusion conditions. Historically, scanning and flash LiDAR systems have been employed in missions ranging from asteroid rendezvous to International Space Station proximity operations \cite{WOODS2016298, rhodes}. More recently, Doppler LiDAR technology has emerged, providing not only range but also line-of-sight velocity information at each measurement point, thus yielding four-dimensional data that enhances state observability \cite{doi:10.2514/6.2025-2432}. These advances make LiDAR an attractive complement to vision-based navigation, where AI corner detections can be fused with LiDAR depth to improve estimation accuracy. In this work, we exploit this synergy by explicitly modeling systematic LiDAR biases and introducing a bias state into the UKF, thereby mitigating discrepancies between projected corner locations and LiDAR point returns and improving the robustness of our measurement model.

\begin{figure}[ht]
    \centering
    \includegraphics[width=\linewidth]{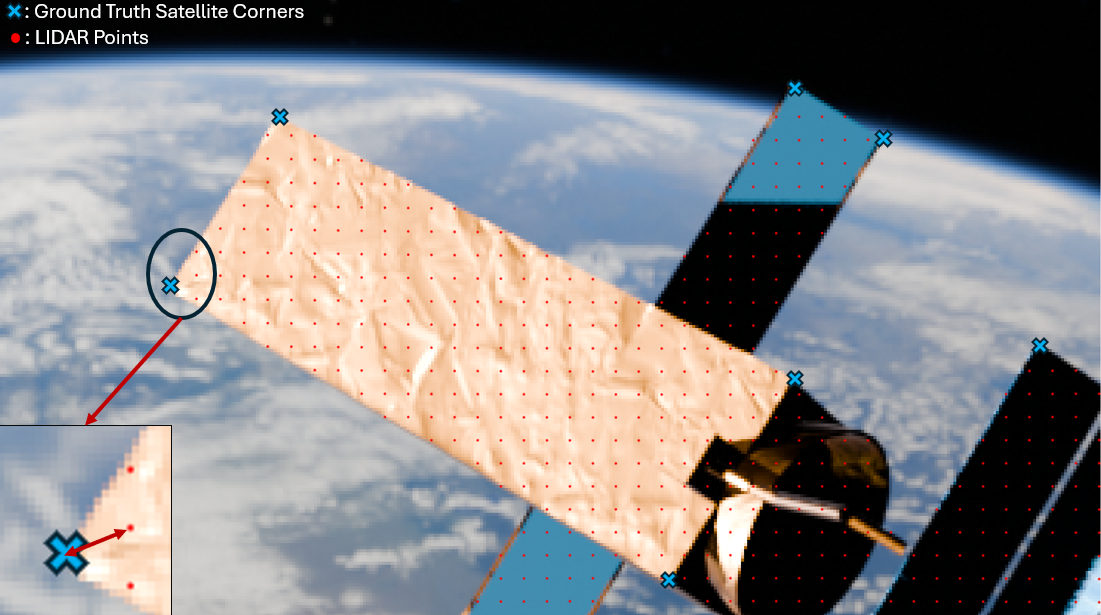}
    \caption{Example frame illustrating the discrepancy between 
    ground-truth satellite corner projections and 
    LiDAR point cloud returns projected into the camera image.}
    \label{fig:lidar_bias}
\end{figure}

To address limitations observed when fusing LiDAR depth with CNN-based corner detections, we introduce an enhanced measurement model that explicitly accounts for systematic discrepancies between projected satellite corners and  LiDAR point returns. In prior results, depth values were generated from simulated ground truth with additive noise, whereas with this modification, we employ a LiDAR-based depth channel. This shift allows us to more realistically capture the measurement characteristics and systematic offsets present in practical operations. Figure~\ref{fig:lidar_bias} highlights the issue, which is that even with perfect ground-truth geometry, nearest-neighbor LiDAR points often fall several pixels away from the true marker locations. This becomes more evident, especially when the chaser is approaching the target from a far distance. To mitigate the resulting bias, we augment the state with a dedicated LiDAR depth bias variable and reformulate the camera–depth measurement model accordingly. The following subsections detail the state augmentation and dynamics, and the revised camera–depth measurement model with bias.

\subsubsection{State Augmentation and Dynamics}
The original state vector is now augmented with a scalar LiDAR depth bias \(b \in \mathbb{R}\) to obtain,
\begin{equation}
\label{eq:aug_state}
\tilde{\mathbf{x}} \;\triangleq\; \begin{bmatrix} \mathbf{x} \\  b \end{bmatrix} \in \mathbb{R}^{13}.
\end{equation}
The bias follows a random walk,
\(
b_{k+1} = b_k + w^b_k,\;\; w^b_k \sim \mathcal{N}(0,\sigma_b^2),
\)
while the \(12\)-dimensional kinematics are propagated with the same translational and rotational dynamics as before, i.e., we do not introduce a deterministic drift for \(b\) in the propagation step. The initial prior and process noise for \(b\) are set as in the code: \(P_{b,0} = 0.2^2\), \(Q_{b} = (10^{-3})^2\).

\subsubsection{Camera-Depth Measurement Model with Bias}
\begin{figure}[H]
    \centering
    \includegraphics[width=\textwidth]{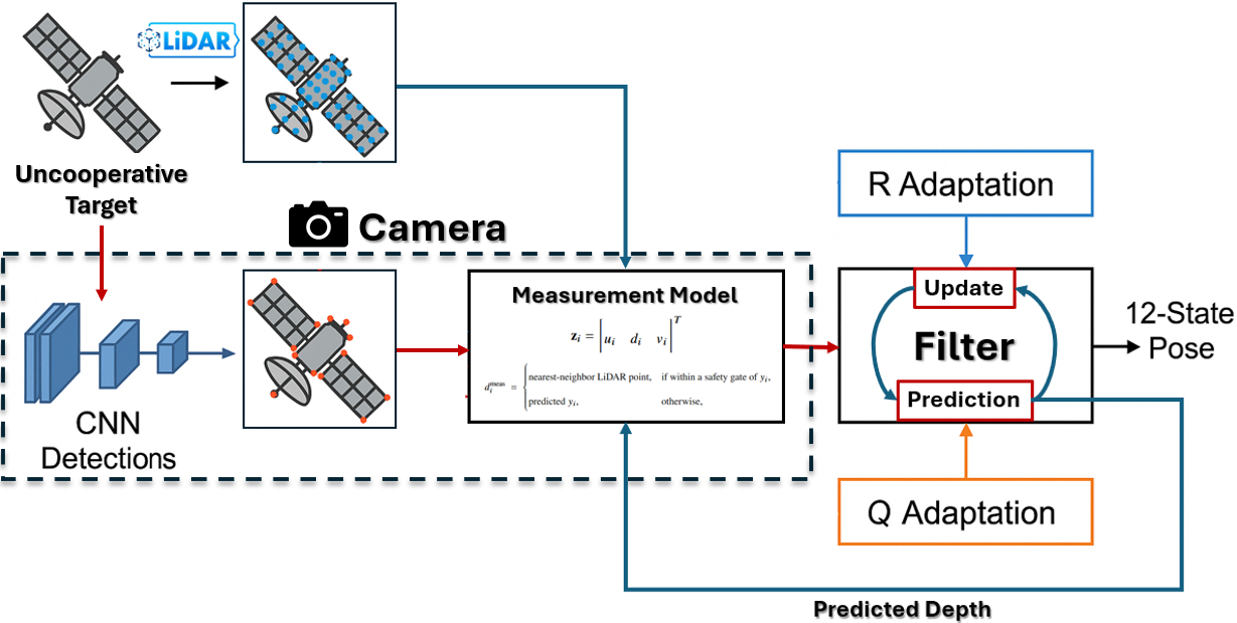}
    \caption{Complete software architecture with filter-camera-LiDAR communication}
    \label{fig:quad}
\end{figure}

\begin{figure}[H]
    \centering    \includegraphics[width=\textwidth]{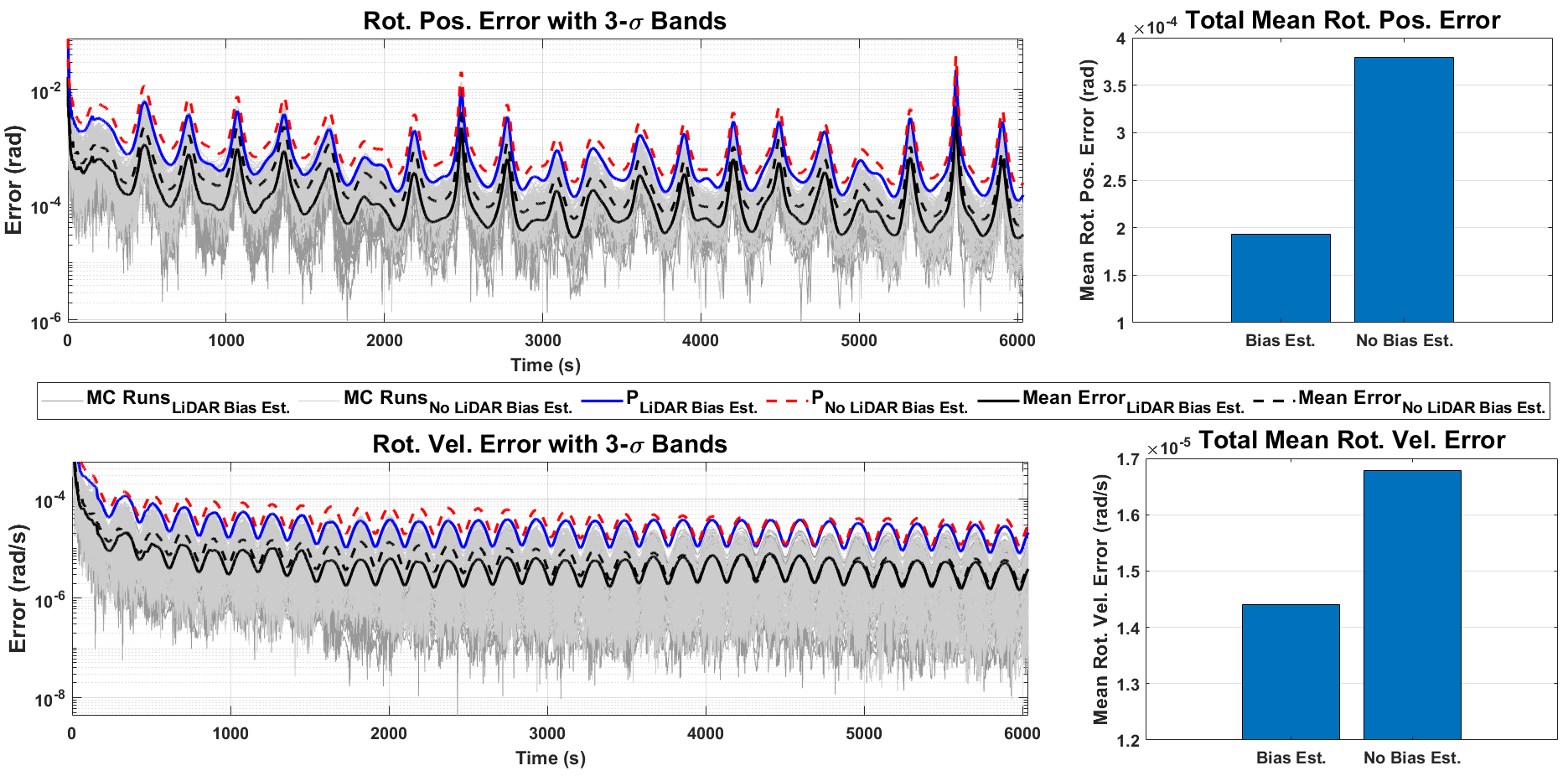}
    \caption{Monte Carlo RMSE comparison for the no-eclipse scenario}
    \label{fig:lidarbest_ne}
\end{figure}

\begin{figure}[H]
    \centering
\includegraphics[width=\textwidth]{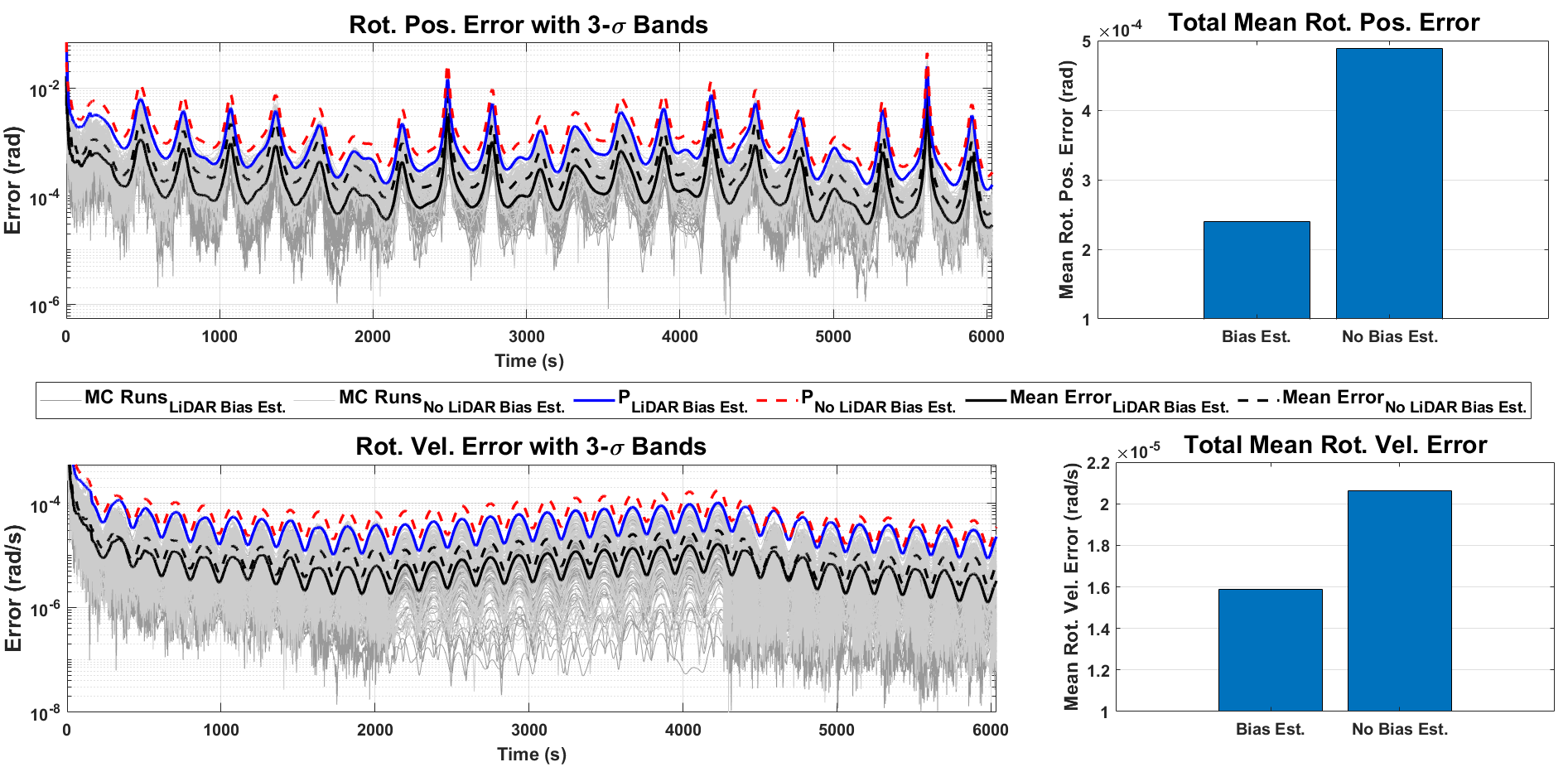}
    \caption{Monte Carlo RMSE comparison for the eclipse scenario}
    \label{fig:lidarbest_e}
\end{figure}

For each visible corner \(i\), in the old model, to get the measurement vector, we projected its body-frame position \(\mathbf{v}_i\) into the camera frame using \(\mathbf{\Gamma}\) and the relative translation \(\mathbf{u}\) defined in Eq. \ref{eq_z}. Pixels ($u_i, v_i$) are obtained with the calibrated intrinsics, and the depth is the second component of the state vector, which is basically the relative translation on $\mathbf{Y}$ axis, additively biased by \(b\).
\begin{equation}
    d_i \;=\; y_i + b.
\end{equation}
Thus each marker contributes a 3-channel RGB-D style measurement \( \mathbf{z}_i = \begin{bmatrix} u_i & d_i & v_i \end{bmatrix}^T \),
\begin{equation}
d_i^{\text{meas}} \;=\; \begin{cases}
\text{nearest-neighbor LiDAR point}, & \text{if within a safety gate of } y_i,\\
\text{predicted } y_i, & \text{otherwise},
\end{cases}
\end{equation}
The safety gate discards LiDAR depths that differ from the predicted \(y_i\) by more than a fixed threshold (defined as 2\,m), limiting catastrophic updates. Collectively, these changes make the filter robust to slowly varying LiDAR bias. So, with this improvement, we have reached the final algorithm framework for this study, and would like to share an appropriate diagram to summarize all components of the algorithm for the readers. Fig.~\ref{fig:quad} visually demonstrates the whole framework in which one can see how each component interacts with the others. To further assess the performance of the updated measurement model, we performed Monte Carlo simulations under two distinct conditions again, one with no eclipse and the other with eclipse periods, and compared these new results with the old ones. For clarity, we present our results for the rotational position and rotational velocity errors, as these are the areas where the most significant improvement is observed. The LiDAR bias correction, along with the updated measurement model, significantly improves the accuracy of the rotational states as shown in the Monte Carlo results in Figures~\ref{fig:lidarbest_ne} and~\ref{fig:lidarbest_e}. 

So, this improvement with the results shown not only makes the whole estimation framework more accurate, but also increases the estimation quality, especially for the rotational states.

\subsection{Filter Consistency}
\begin{figure}[H]
    \centering
\includegraphics[width=0.95\textwidth]{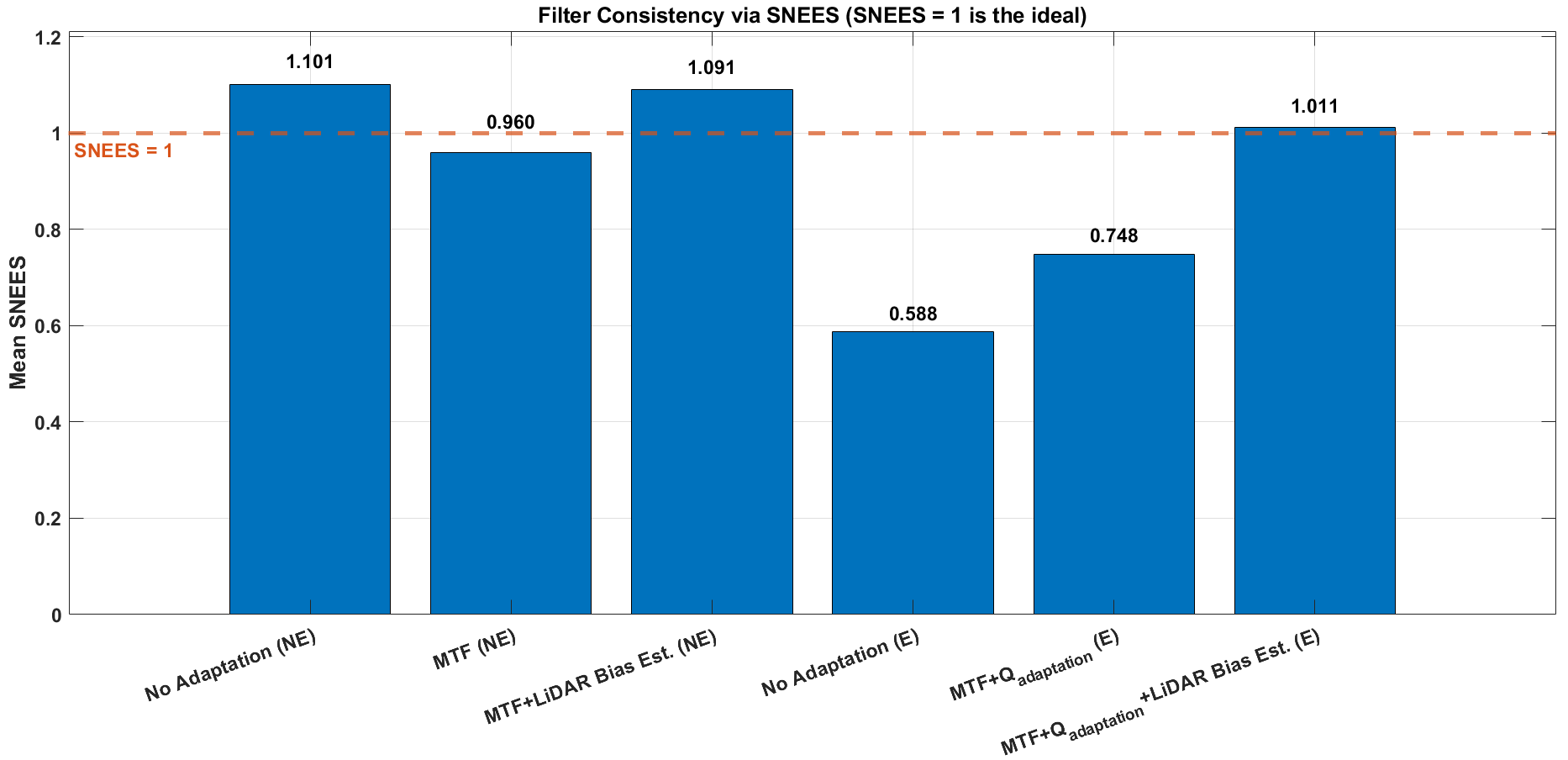}
    \caption{SNEES comparison for filter consistency of each method}
    \label{fig:snees_bar}
\end{figure}

Moreover, in order to assess filter consistency, our work adopts the Scaled Normalized Estimation Error Squared (SNEES) metric, a commonly used statistical indicator that evaluates the alignment between the predicted uncertainty and the actual estimation error. As formulated and shown in \cite{GiraldoGrueso2025Optimal}, SNEES compares the squared estimation error with the predicted covariance matrix across Monte Carlo trials. A SNEES value close to one indicates a statistically consistent filter, while significant deviations suggest overconfidence (SNEES $>$ 1) or excessive conservatism (SNEES $<$ 1).
\begin{equation}
\mathrm{SNEES}(k) = \frac{1}{N_s N_m} \sum_{j=1}^{N_m} 
\left( \mathbf{x}_{k,j} - \hat{\mathbf{x}}_{k|k,j} \right)^T 
\mathbf{P}_{k|k,j}^{-1} 
\left( \mathbf{x}_{k,j} - \hat{\mathbf{x}}_{k|k,j} \right)
\end{equation}
Here, \( \mathbf{x}_{k,j} \) represents the true state vector at time step \( k \) for the \( j \)-th Monte Carlo run, and \( \hat{\mathbf{x}}_{k|k,j} \) denotes the corresponding filter estimate. The term \( \mathbf{P}_{k|k,j} \) is the estimated error covariance matrix by UKF. The constants \( N_s \) and \( N_m \) refer to the state dimension and the total number of Monte Carlo runs (e.g., \( N_m = 100 \)), respectively. We compute SNEES using the updated covariance and the Monte Carlo sample errors for estimated states, leveraging our existing filter outputs without introducing additional simulation overhead. This approach allows us to quantify how well the filter captures the true uncertainty during conditions of measurement degradation, such as eclipses or marker dropout, and further substantiates the robustness of our dual-adaptive framework. Conceptually, the SNESS value of the filter indicates how well the filter estimates its own accuracy level. Thus, to evaluate the consistency of all the proposed methods, we analyzed the SNEES over time for both the adaptive and non-adaptive cases as shown in Fig.~\ref{fig:snees_bar}.

The SNEES analysis highlights the impact of our adaptive strategies on filter consistency. Both \textbf{MTF} only and $\textbf{MTF}+\textbf{Q}_{adaptation}$ reduce the gap between the predicted covariance and the actual estimation error, with mean SNEES values close to one (0.960 for \textbf{MTF} in the non-eclipse case and 0.748 for $\textbf{MTF}+\textbf{Q}_{adaptation}$ in eclipse). This indicates that the adaptive schemes successfully balance confidence with uncertainty, avoiding the overconfidence and overconservatism seen in the non-adaptive baselines (1.101 in non-eclipse and 0.588 in eclipse). However, incorporating LiDAR bias estimation in both cases, the filter tends toward better consistency (1.091 with \textbf{MTF}+LiDAR bias estimation, and 1.011 with $\textbf{MTF}+\textbf{Q}_{adaptation}$). These results confirm that adaptation mechanisms substantially enhance consistency, while also revealing that coupling LiDAR bias estimation makes the filter more consistent, while also reducing the overall error, especially for rotational states. In summary, the dual-adaptive framework provides robust improvements in filter consistency across scenarios.

\section{Conclusions}
\label{sec:conclusions}
This work presents a robust and dynamically adaptive UKF framework tailored for vision-based relative navigation under challenging conditions such as intermittent marker visibility and eclipses. By leveraging adaptive methodologies, we propose a dual adaptation strategy that adjusts both the process noise covariance $\mathbf{Q}$ and the measurement noise covariance $\mathbf{R}$ in response to observed filter behavior. Our adaptation of $\mathbf{R}$ is guided by a residual-consistency formulation that filters out non-Gaussian measurement deviations. On the other hand, the adjustment of $\mathbf{Q}$ is driven by state divergence metrics during occlusion periods, inspired by forward prediction mismatch and smoothed uncertainty propagation. Results demonstrate that our method significantly improves estimation accuracy, compared to static-filter configurations and prior techniques. Together, they yield significant reductions in estimation error across all 12 state variables compared to non-adaptive filters, and consistently lower SNEES values, particularly during challenging eclipse phases. Notably, we outperform recent neural-filter-based efforts such as the approach in \cite{candan}, especially in eclipse phases where marker observations are sparse or fully absent. Our filter maintains stability and bounded covariance even during complete visual dropout, while previous approaches suffer from overly confident covariances and larger, fluctuating mean errors. Furthermore, these methods do not require manual tuning and generalize across dynamic lighting and visibility conditions, positioning them as practical candidates for onboard deployment. The framework's real-world applicability was further enhanced by incorporating a LiDAR sensor model and explicitly estimating its measurement bias, improving overall accuracy, especially for rotational states

As a future work, our immediate priority is to transition from high-fidelity simulation to hardware-in-the-loop testing. A dedicated laboratory is being established in the Department of Aerospace Engineering at Iowa State University to recreate more realistic space-borne sensing conditions. This facility will integrate motion platforms, camera and LiDAR sensors, and controlled lighting environments to emulate on-orbit dynamics, enabling experimental validation of the adaptive UKF under physical constraints. Together, these directions pave the way toward robust, flight-ready implementations of adaptive filtering for autonomous proximity operations. Another important direction is to extend the framework from relative navigation to absolute state estimation, allowing autonomous spacecraft to perform both cooperative and uncooperative operations with broader mission applicability. Together, these directions pave the way toward robust, flight-ready implementations of adaptive filtering for autonomous proximity operations.

Looking further ahead, the framework can be extended to tackle more complex and ambitious scenarios. One major research thrust involves moving beyond known targets like ENVISAT to address unknown, non-cooperative objects, effectively evolving the problem into the domain of Simultaneous Localization and Mapping (SLAM). The filter could also be augmented to estimate key physical properties of the target, such as its inertia matrix, which is vital for capture planning. Ultimately, the goal is to transform the system from a passive estimator into an active navigator, where the filter's uncertainty predictions are used to drive an information-theoretic guidance law that optimizes the chaser's trajectory for the most rapid and accurate pose estimation. These advancements will continue to push the boundaries of autonomous navigation, enabling safer and more capable proximity operations in complex space environments.

\section*{Declarations}

\begin{itemize}
\item The authors declare that no funds, grants, or other support were received during the preparation of this manuscript.
\end{itemize}

\begin{appendices}




\end{appendices}


\bibliography{sn-bibliography}

\end{document}